\documentclass[10pt,twocolumn,letterpaper]{article}

\usepackage[pagenumbers]{cvpr} 

\usepackage{amsthm}
\newtheorem{theorem}{Theorem}
\newtheorem{assumption}{Assumption}

\usepackage{algorithm}
\usepackage{algorithmic}
\usepackage{multirow}
\usepackage{booktabs}
\usepackage{makecell}
\usepackage[table]{xcolor}
\usepackage{tabularx}
\definecolor{fidbase}{RGB}{88, 140, 125} 
\definecolor{clipbase}{RGB}{155, 126, 183}

\definecolor{cvprblue}{rgb}{0.21,0.49,0.74}
\usepackage[pagebackref,breaklinks,colorlinks,allcolors=cvprblue]{hyperref}

\def\paperID{11529} 
\def\confName{CVPR}
\def\confYear{2026}

\title{CFG-Ctrl: Control-Based Classifier-Free Diffusion Guidance}

\author{
Hanyang Wang\footnotemark[1],
Yiyang Liu\footnotemark[1],
Jiawei Chi,
Fangfu Liu,
Ran Xue,
Yueqi Duan\footnotemark[2]\\
Tsinghua University
}

\begin{document}
\maketitle

\renewcommand{\thefootnote}{\fnsymbol{footnote}}

\footnotetext{$^*$Equal contribution. $^\dagger$Corresponding author.}

\renewcommand{\thefootnote}{\arabic{footnote}}

\begin{abstract}
Classifier-Free Guidance (CFG) has emerged as a central approach for enhancing semantic alignment in flow-based diffusion models. In this paper, we explore a unified framework called \textbf{CFG-Ctrl}, which reinterprets CFG as a control applied to the first-order continuous-time generative flow, using the conditional-unconditional discrepancy as an error signal to adjust the velocity field. From this perspective, we summarize vanilla CFG as a proportional controller (P-control) with fixed gain, and typical follow-up variants develop extended control-law designs derived from it. However, existing methods mainly rely on linear control, inherently leading to instability, overshooting, and degraded semantic fidelity especially on large guidance scales. To address this, we introduce Sliding Mode Control CFG (\textbf{SMC-CFG}), which enforces the generative flow toward a rapidly convergent sliding manifold. Specifically, we define an exponential sliding mode surface over the semantic prediction error and introduce a switching control term to establish nonlinear feedback-guided correction. Moreover, we provide a Lyapunov stability analysis to theoretically support finite-time convergence. Experiments across text-to-image generation models including Stable Diffusion 3.5, Flux, and Qwen-Image demonstrate that SMC-CFG outperforms standard CFG in semantic alignment and enhances robustness across a wide range of guidance scales. Project Page: \url{https://hanyang-21.github.io/CFG-Ctrl}.
\end{abstract}    
\section{Introduction}
\label{sec:intro}

\begin{figure}
    \centering
    \includegraphics[width=\linewidth]{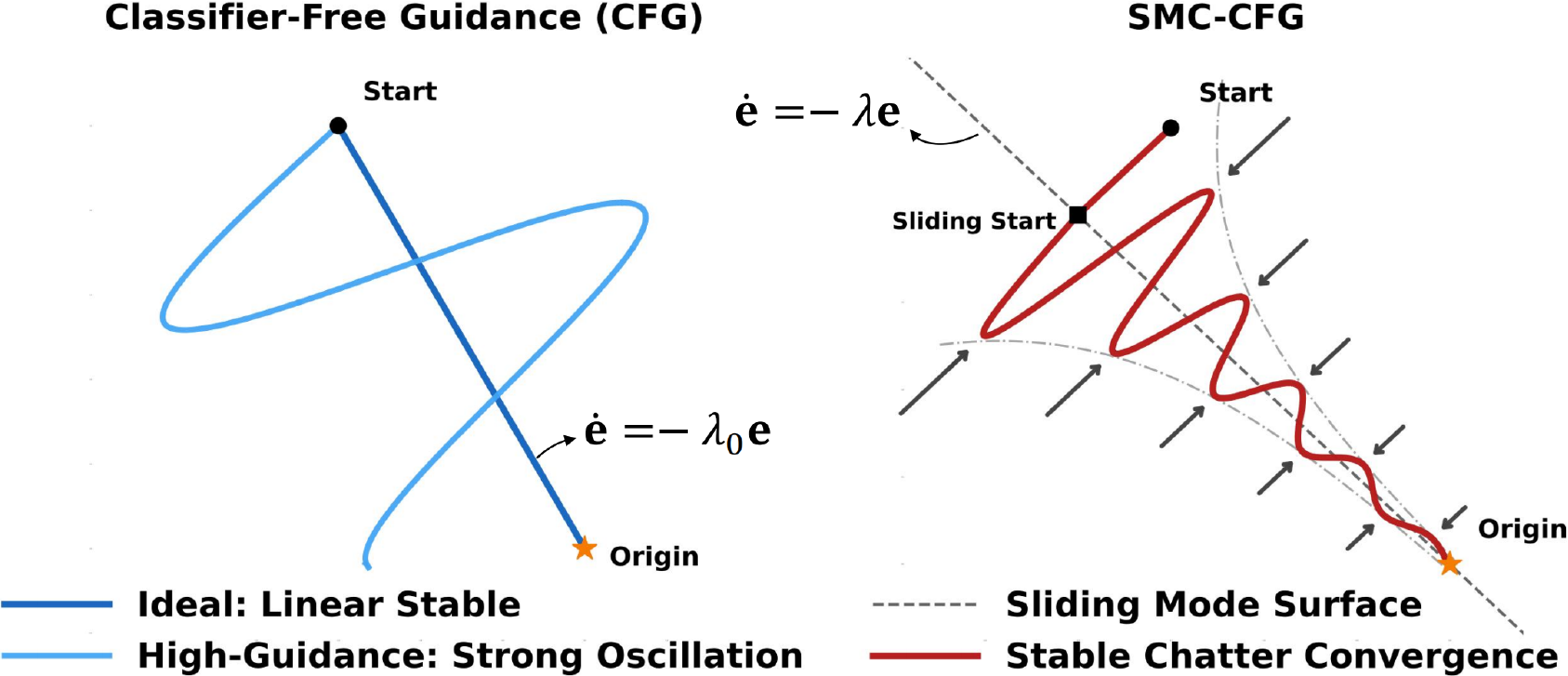}
    \vspace{-5mm}
    \caption{\textbf{Phase diagram in the $\mathbf{e}$-$\dot{\mathbf{e}}$ plane.} We schematically illustrate the convergence patterns of CFG and the proposed SMC-CFG. \textit{Left}: CFG's ideal linear convergence trajectory and the strong oscillatory divergence under high guidance scales. \textit{Right}: the proposed SMC-CFG, through a switching-forcing mechanism, drives the system states toward the sliding mode surface governed by parameter $\lambda$, achieving robust and rapid convergence.}
    \vspace{-5mm}
    \label{fig:convergence}
\end{figure}

Diffusion models~\cite{ho2020ddpm,song2020score,song2020ddim} have recently achieved state-of-the-art performance in high-fidelity image synthesis across diverse domains~\cite{rombach2022ldm,peebles2023dit}. Building on the similar probabilistic formulation, flow matching~\cite{liu2022flow,lipman2022flow_matching} provides a more straightforward alternative by directly estimating deterministic velocity fields, realizing stable training and faster sampling than diffusion~\cite{grathwohl2018ffjord,fan2025cfg-zero}. These flow-based methods have demonstrated strong capability across text-to-image~\cite{flux2024,esser2024sd3}, text-to-video~\cite{yang2024cogvideox,wan2025wan,kong2024hunyuanvideo}, and other visual generation applications~\cite{zhao2025hunyuan3d,labs2025flux1kontextflowmatching,wang2025videoscene}.

A key technique widely adopted in diffusion models is Classifier-Free Guidance (CFG)~\cite{ho2022cfg}, which enhances semantic alignment between the generated sample and the input condition. Previous studies commonly interpret CFG as a linear extrapolation between unconditional and conditional predictions within deterministic diffusion flows~\cite{saini2025reccfg++}. While this perspective offers an intuitive interpretation, the resulting linear extrapolation can distort the generative trajectory from the learned data manifold, leading to oversaturated colors, warped structures, and strong sensitivity to the guidance scale~\cite{chung2024cfg++}. Several improved methods have been proposed to alleviate these issues, including linear recomposition~\cite{xia2025rectifiedcfg}, orthogonal decomposition~\cite{sadat2024APG,fan2025cfg-zero}, and dynamic weighting schedules~\cite{wang2024analysis,saini2025reccfg++}.

We observe that the discrepancy between the conditional and unconditional velocity predictions gradually diminishes in diffusion flow progress, effectively serving as a natural error signal. This observation motivates us to reinterpret CFG not as a static extrapolation rule, but as a form of feedback control applied to the latent generative flow. Based on this observation, we explore a unified theoretical framework called \textbf{CFG-Ctrl} for Classifier-Free Guidance in flow matching diffusion. Under this CFG-Ctrl paradigm, the standard CFG corresponds to a proportional controller (P-control) that amplifies the semantic error with a fixed gain and feeds it back into the system, while existing CFG variants can be regarded as alternative designs of feedback control laws. However, most of these methods rely on approximately linear control laws for feedback, which cannot ensure stable convergence when the underlying generative dynamics become highly nonlinear—particularly as model capacity increases or the guidance scale becomes large as shown in Figure~\ref{fig:convergence} (left).

To address this, we further propose Sliding Mode Control CFG (\textbf{SMC-CFG}), a control-based guidance mechanism that directs the flow trajectory onto a rapidly converging sliding mode surface. This design draws on the proven success of Sliding Mode Control (SMC)~\cite{edwards1998sliding,zeinali2010adaptive} in stabilizing nonlinear dynamical systems. As shown in Figure~\ref{fig:convergence} (right), our approach constructs a sliding mode surface over the semantic prediction error, corresponding to the gray dashed line in the figure. We also introduce a switching control term that enforces nonlinear, feedback-driven corrective force, which are represented by the arrows at both sides of the convergence curve. This design adaptively regulates the evolution of the flow trajectory and preserves stability even under strong guidance. To theoretically substantiate convergence, we provide a Lyapunov stability analysis based on the principle of monotonically decreasing energy, demonstrating that SMC-CFG supports finite-time convergence toward the desired semantic manifold. Extensive experiments on three state-of-the-art text-to-image (T2I) models show that SMC-CFG consistently improves semantic fidelity, reduces visual artifacts, and maintains robustness across multiple semantic and perceptual metrics. Our contributions are summarized as follows:
\begin{itemize}
    \item We explore CFG-Ctrl, a novel theoretical framework for Classifier-Free Guidance in flow matching models grounded in control theory, unifying the systematic interpretation of diverse guidance strategies.
    \item We propose SMC-CFG, a sliding-mode-based nonlinear feedback controller for flow models, and prove finite-time convergence under Lyapunov stability analysis.
    \item Extensive experiments across multiple diffusion backbones demonstrate that SMC-CFG achieves superior semantic fidelity, visual coherence, and robustness, particularly under high guidance scales.
\end{itemize}

\section{Related Work}
\label{sec:related_work}
\textbf{Diffusion and Flow Matching.}
Diffusion models~\cite{ho2020ddpm, song2020ddim, song2020score} have garnered significant attention in recent years as a class of generative models that iteratively transform simple distributions into more complex ones, ultimately generating high-quality samples. Early diffusion models define a forward diffusion process, where noise is gradually added to data samples, typically starting from a simple prior such as an isotropic Gaussian. The reverse process is then learned by training a neural network to estimate the score function of the data distribution~\cite{rombach2022ldm,song2020score}, enabling the model to progressively recover the original data. More recently, flow matching~\cite{liu2022flow, lipman2022flow_matching} has been proposed to model the transformation process via a learned velocity field, which simplifies the generative formulation and leads to better empirical performance. This paradigm has been widely adopted in large-scale foundation models across multiple domains, including image generation~\cite{flux2024,wu2024unique3d,wu2025qwen-image}, video generation~\cite{hacohen2024ltx,liu2024physics3d,liu2025video-t1}, and 3D content generation~\cite{liu2026reconx,yao2026anchoreddream,liu2025dreamreward}, demonstrating its scalability and strong performance advantages.

\noindent\textbf{Guidance in Diffusion.}
Guidance techniques play a crucial role across a wide range of visual tasks~\cite{liu2025langscene,yao2025airroom}. In diffusion-based generative models, guidance emerges as a core mechanism for conditional generation. Early approaches such as Classifier Guidance (CG)~\cite{dhariwal2021diffusion} improve sample quality by leveraging an external classifier to steer the denoising process toward desired semantic targets, but require training a separate noise-aware classifier and are difficult to scale to complex or multimodal conditioning signals. To address these limitations, Classifier-Free Guidance (CFG)~\cite{ho2022cfg} was introduced, enabling conditional generation~\cite{saharia2022photorealistic,ruiz2023dreambooth,liu2024make} without an auxiliary classifier. By jointly training the diffusion model with and without conditioning inputs, CFG allows flexible control at inference time through a simple interpolation between conditional and unconditional predictions. Subsequent works~\cite{zheng2024characteristic,kynkaanniemi2024applying,lin2024common,chung2024cfg++} explore adaptive guidance strategies for CFG by dynamically adjusting the guidance scale~\cite{wang2024analysis,xia2025rectifiedcfg} or refining the guidance direction~\cite{sadat2024APG} to mitigate oversaturation, thereby improving fidelity and reducing artifacts. Building upon these, recent studies~\cite{fan2025cfg-zero, saini2025reccfg++} have extended the CFG to flow matching models. For example, CFG-Zero$^\star$~\cite{fan2025cfg-zero} introduces an optimized guidance scale to correct velocity estimation, while Rectified-CFG++~\cite{saini2025reccfg++} proposes an adaptive predictor–corrector scheme that integrates the deterministic efficiency of rectified flows. These methods demonstrate that guidance remains a powerful and extensible mechanism for controllable and efficient generative modeling.

\noindent \textbf{Control Theory.}
Control theory provides a foundational framework for designing systems that can regulate their behavior to achieve desired objectives. Its principles are fundamental to ensuring the performance, safety, and efficiency of complex systems across aerospace~\cite{beard2012small, bryson2018applied}, robotics~\cite{siciliano2009robotics, yoshikawa1990foundations}, and industrial process control~\cite{seborg2016process, mayne2000constrained}. Among various approaches, Proportional–Integral–Derivative (PID) control~\cite{1995pid} remains one of the most widely adopted strategies due to its simplicity and broad applicability, effectively balancing responsiveness, stability, and steady-state accuracy using feedback errors.
Beyond PID, advanced paradigms address more complex challenges: Model Predictive Control (MPC) optimizes future actions based on a system model~\cite{garciaModelPredictiveControl1989}, while Adaptive Control adjusts parameters online to manage uncertainties~\cite{astromAdaptiveControl1991}. Furthermore, robust control strategies guarantee stability against defined model inaccuracies. Sliding Mode Control (SMC)~\cite{edwards1998sliding}, as a prominent example of robust control, introduces a discontinuous law that forces the system trajectory onto a predefined manifold, ensuring exceptional resilience to disturbances. These diverse control strategies have inspired recent efforts to integrate feedback-based and stability-driven principles into learning-based and generative modeling frameworks.

\section{Method}

\subsection{Preliminaries}
\label{sec:preliminaries}
\textbf{Classifier-Free Guidance (CFG)}~\cite{ho2022cfg} introduces guidance by linearly interpolating between the conditional and unconditional velocity fields. Let $\mathbf{v}_\theta(\mathbf{x}_t, t, \varnothing)$ denote the \emph{unconditional} velocity, obtained by dropping the condition $\mathbf{c}$ during training. The guided velocity is computed as:
\begin{equation}
    \label{eq:cfg1}
    \resizebox{0.9\hsize}{!}{$\hat{\mathbf{v}}_\theta(\mathbf{x}_t, t, \mathbf{c})
    = \mathbf{v}_\theta(\mathbf{x}_t, t, \varnothing)
    + w \cdot \bigl( \mathbf{v}_\theta(\mathbf{x}_t, t, \mathbf{c})
    - \mathbf{v}_\theta(\mathbf{x}_t, t, \varnothing) \bigr),$}
\end{equation}
where $w \ge 1$ is the guidance weight. Rearranging yields:
\begin{equation}
    \label{eq:cfg2}\hat{\mathbf{v}}_\theta(\mathbf{x}_t, t, \mathbf{c})
    = (1-w)\,\mathbf{v}_\theta(\mathbf{x}_t, t, \varnothing)
    + w\,\mathbf{v}_\theta(\mathbf{x}_t, t, \mathbf{c}).
\end{equation}
When $w = 1$, the model reduces to the standard conditional predictor. Increasing $w > 1$ amplifies the conditional component, improving semantic alignment at the cost of reduced sample diversity.

\noindent\textbf{Weight-Scheduler}~\cite{wang2024analysis} introduces a time-varying guidance weight $w(t)$ in place of the fixed weight $w$ in standard CFG. The guided velocity becomes:
\begin{equation}
\label{eq:weigt-scheduler}
\resizebox{0.9\hsize}{!}{$
    \hat{\mathbf{v}}_\theta(\mathbf{x}_t, t, \mathbf{c})
    = \mathbf{v}_\theta(\mathbf{x}_t, t, \varnothing)
    + w(t) \cdot \bigl( \mathbf{v}_\theta(\mathbf{x}_t, t, \mathbf{c})
    - \mathbf{v}_\theta(\mathbf{x}_t, t, \varnothing) \bigr).
    $}
\end{equation}
Here, the scheduler $w(t)$ is a monotonically increasing function of the denoising step to avoid overshooting the guidance in the initial stages.

\noindent\textbf{Adaptive Projected Guidance (APG)}~\cite{sadat2024APG} mitigates oversaturation by down-weighting the component of the guidance direction that is parallel to the conditional prediction. The standard CFG update direction
$\Delta \mathbf{v}_t = \mathbf{v}_\theta(\mathbf{x}_t, t, \mathbf{c}) - \mathbf{v}_\theta(\mathbf{x}_t, t, \varnothing)$
is decomposed into parallel and orthogonal components:
\begin{equation}
\label{eq:apg-projection}
\resizebox{0.9\hsize}{!}{$
    \Delta \mathbf{v}_t^{\parallel}
    = \frac{\langle \Delta \mathbf{v}_t,\, \mathbf{v}_\theta(\mathbf{x}_t, t, \mathbf{c}) \rangle}
           {\|\mathbf{v}_\theta(\mathbf{x}_t, t, \mathbf{c})\|^2}
      \, \mathbf{v}_\theta(\mathbf{x}_t, t, \mathbf{c}), \quad
    \Delta \mathbf{v}_t^{\perp}
    = \Delta \mathbf{v}_t - \Delta \mathbf{v}_t^{\parallel}.$}
\end{equation}
APG reduces oversaturation by down-weighting the parallel term. The guided velocity then becomes:
\begin{equation}
\label{eq:apg}
\resizebox{0.9\hsize}{!}{$
    \mathbf{v}_{\text{APG}}(\mathbf{x}_t, t, \mathbf{c})
    = \mathbf{v}_\theta(\mathbf{x}_t, t, \varnothing)
    + w \cdot\bigl(\Delta \mathbf{v}_t^{\perp} + \eta\,\Delta \mathbf{v}_t^{\parallel}\bigr),
    \quad \eta \le 1.$}
\end{equation}
APG chooses $\eta<1$ to suppress oversaturation while preserving the quality-enhancing orthogonal component.

\subsection{Motivation}
\label{sec:motivation}
Classifier-Free Guidance (CFG) has demonstrated remarkable empirical success across numerous diffusion-based generative models and related applications. In prior formulations, CFG can be viewed as a linear extrapolation within deterministic diffusion flows~\cite{saini2025reccfg++}, as expressed in Eq.~\eqref{eq:cfg1}. We denote the guidance term as
\begin{equation}
\mathbf{e}(t)=\mathbf{v}_\theta(\mathbf{x}_t, t, \mathbf{c})- \mathbf{v}_\theta(\mathbf{x}_t, t, \varnothing).
\label{eq:error_definition}
\end{equation}

Ideally, during the denoising process from time step $T$ to $0$, CFG continuously injects conditional information into the trajectory $\mathbf{x}_t$. This mechanism progressively enriches the semantic content encoded in $\mathbf{x}_t$ as the timestep decreases. In the final stages of denoising, when most semantic information has already been embedded in $\mathbf{x}_t$, the conditional and unconditional predictions tend to converge, \ie, $\mathbf{v}_\theta(\mathbf{x}_t, t, \mathbf{c}) \approx \mathbf{v}_\theta(\mathbf{x}_t, t, \varnothing)$, such that both $\mathbf{e}$ and its temporal derivative $\dot{\mathbf{e}}$ approach zero.
This ideal behavior can be viewed geometrically as a guidance process evolving on the $\left (\mathbf{e}, \dot{\mathbf{e}}\right )$ plane, aiming to drive the system state toward the equilibrium point $(0,0)$. The most direct and stable convergence path under such a setting corresponds to the first-order linear system:
\begin{equation}
\dot{\mathbf{e}} = -\lambda_0 \cdot \mathbf{e}, \quad \lambda_0 = -\frac{\dot{\mathbf{e}}(T) \cdot \mathbf{e}(T)}{\|\mathbf{e}(T)\|^2}, \quad (\mathbf{e}(T) \neq \mathbf{0}).
\end{equation}
From the viewpoint of differential-equation, it implies a rapid and stable exponential convergence.

In practice, however, the assumption of collinear $(\mathbf{e}, \dot{\mathbf{e}})$ relationship rarely holds, especially when model capacity increases and the CFG guidance scale is enlarged. The resulting system becomes highly nonlinear, and the standard CFG formulation may exhibit oscillatory or divergent behavior, as illustrated in Fig.~\ref{fig:convergence} (left). Such instability often manifests as color distortion, loss of fine details, or inconsistent textures in generated images~\cite{chung2024cfg++,zheng2024characteristic,sadat2024APG}.

Motivated by the effectiveness of control methods in stabilizing oscillatory behavior and ensuring convergence in dynamical systems, we revisit CFG from a control-theoretic perspective. Rather than treating CFG as a static extrapolation method, we propose to view CFG as a feedback control process that actively regulates the evolution of $\mathbf{e}(t)$, driving it toward the equilibrium in a principled, rate-aware manner.

\begin{table*}[h]
\centering
\caption{\textbf{Typical CFG variants under CFG-Ctrl formulation.} We summarize the key components of various methods under the control formulation, along with their corresponding types of control interpretations.}
\vspace{-3mm}
    \resizebox{\linewidth}{!}{
    \begin{tabular}{lcccc}
        \toprule
        \textbf{Method} & \textbf{Gain $K_t$} & \textbf{Operator $\Pi_t$} & \textbf{Error $\mathbf{e}(t)$} & \textbf{Control Interpretation}\\
        \midrule
        CFG~\cite{ho2022cfg} & $w$ & $I$ & $\Delta \mathbf{v}_\theta(t)$ & Proportional control \\[10pt]
        
        Weight Scheduler~\cite{wang2024analysis} & $w(t)$ & $I$ & $\Delta \mathbf{v}_\theta(t)$ & Time-varying gain scheduling \\[10pt]
        
        APG~\cite{sadat2024APG} & $w\begin{bmatrix} I\;\; \eta I \end{bmatrix}$ & $\begin{bmatrix} I - P_t \\ P_t \end{bmatrix}, P_t = \frac{\mathbf{v}_\theta(\mathbf{c})\mathbf{v}_\theta(\mathbf{c})^\top}{|\mathbf{v}_\theta(\mathbf{c})|^2}$ & $\Delta \mathbf{v}_\theta(t)$ & Projection-based Feedback Control \\[10pt]
        
        CFG-Zero$^\star$~\cite{fan2025cfg-zero} & $\begin{bmatrix} wI\;\; \frac{s_t}{1-s_t}I \end{bmatrix}$, $s_t=\frac{\mathbf{v}_\theta(\mathbf{c})^\top\mathbf{v}_\theta(\mathbf{\varnothing})}{|\mathbf{v}_\theta(\mathbf{\varnothing})|^2}$ & $\begin{bmatrix} I - P_t \\ P_t \end{bmatrix}, P_t = \frac{\mathbf{v}_\theta(\mathbf{\varnothing})\mathbf{v}_\theta(\mathbf{\varnothing})^\top}{|\mathbf{v}_\theta(\mathbf{\varnothing})|^2}$ & $\Delta \mathbf{v}_\theta(t)$ & Projection-based Feedback Control \\[10pt]
        
        Rectified-CFG++~\cite{saini2025reccfg++} & $\begin{bmatrix}
            I \;\; \alpha(t)I
        \end{bmatrix}, \alpha(t)=\lambda_{max}(1-t)^\gamma$ & $I$ & $\begin{bmatrix}
            \Delta \mathbf{v}_\theta(t) \\
            \Delta \mathbf{v}_\theta(t-\frac{\Delta t}{2})
        \end{bmatrix}$ & Model Predictive Control \\[10pt]
        
        SMC-CFG & $w$ & $I$ & $\Delta \mathbf{v}_\theta(t)-k \cdot \text{sign}(\mathbf{s}_t)$ & Sliding Mode Control \\
        \bottomrule
    \end{tabular}
    }
\vspace{-3mm}
\label{tab:cfg-family}
\end{table*}

\subsection{Theoretical Formulation of CFG-Ctrl}
\label{sec:formulation}

In this section, we introduce CFG-Ctrl, a unified theoretical framework for CFG in flow matching models, which systematically interprets diverse guidance strategies. We first model the flow matching sampling process as a continuous-time controlled dynamical system. Let $\mathbf{x}_t \in \mathcal{X}\subseteq \mathbb{R}^d$ denote the latent state at time $t \in [0, T]$, whose evolution follows the control-affine ordinary differential equation
\begin{equation}
    \frac{d\mathbf{x}_t}{dt} = \mathbf{v}_\theta(\mathbf{x}_t,t) + \mathbf{G}(\mathbf{x}_t,t)\mathbf{u}_t,
    \label{eq:general-feedback}
\end{equation}
with the initial condition $\mathbf{x}_0 \sim \mathcal{N}(0,\mathbf{I})$. In Eq.~\eqref{eq:general-feedback}, $\mathbf{v}_\theta: \mathcal{X} \times [0, T] \to \mathcal{X}$ is the pre-trained velocity field, $\mathbf{G}: \mathcal{X} \times [0,T]\to \mathbb{R}^{d\times m}$ is the input mapping matrix, and $\mathbf{u}_t \in \mathcal{U} \subseteq \mathbb{R}^m$ is the guidance control input.

Guidance mechanisms act directly in the latent coordinates without cross-space transformations; hence we set $\mathbf{G}(\mathbf{x}_t,t)=\mathbf{I}_d$ (full actuation, $m=d$), reducing Eq.~\eqref{eq:general-feedback} to the additive-velocity form
\begin{equation}
    \frac{d\mathbf{x}_t}{dt} = \mathbf{v}_\theta(\mathbf{x}_t, t) + \mathbf{u}_t.
    \label{eq:additive-velocity}
\end{equation}

To better analyze guidance mechanisms, we propose to formulate the control signal $\mathbf{u}_t$ using a general state-feedback law, decomposing it into two key components:
\begin{equation}
     \mathbf{u}_t = K_t\, \Pi_t\!\big(\mathbf{e}(t)\big).
     \label{eq:feedback-law}
\end{equation}
Here, $\mathbf{e}(t)$, defined in Eq.~\eqref{eq:error_definition}, is regarded as the semantic error of the system. We term $K_t$ the guidance schedule, as it schedules the guidance strength, and $\Pi_t$ the direction operator, as it shapes the correction direction (\eg, normalization or projection).

Under the CFG-Ctrl formulation, we can interpret the standard CFG as a specific, simple instance of this general control law. The standard CFG update Eq.~\eqref{eq:cfg1} modifies the closed-loop dynamics as:
\begin{equation}
\begin{aligned}
     \frac{d\mathbf{x}_t}{dt}
     &= \mathbf{v}_\theta(\mathbf{x}_t,t, \varnothing) 
     + w\left(\mathbf{v}_\theta(\mathbf{x}_t,t,\mathbf{c})
     - \mathbf{v}_\theta(\mathbf{x}_t,t,\varnothing)\right),
\end{aligned}
\label{eq:cfg-feedback}
\end{equation}
where $w$ is the guidance scale. This specific form is recovered when the guidance schedule $K_t$ is a constant scalar and the direction operator $\Pi_t$ is identity:
\begin{equation}
     K_t = w, \qquad \Pi_t = I.
\end{equation}
Substituting these into the closed-loop dynamics $\frac{d\mathbf{x}_t}{dt} = \mathbf{v}_\theta(\mathbf{x}_t,t, \varnothing) + K_t\,\Pi_t(\mathbf{e}(t))$ in Eq.~\eqref{eq:additive-velocity} and~\eqref{eq:feedback-law} yields:
\begin{equation}
\begin{aligned}
     \frac{d\mathbf{x}_t}{dt}
     &= \mathbf{v}_\theta(\mathbf{x}_t,t, \varnothing) + K_t\,\Pi_t(\mathbf{e}(t)) \\
     &= \mathbf{v}_\theta(\mathbf{x}_t,t, \varnothing) 
     + w\left(\mathbf{v}_\theta(\mathbf{x}_t,t,\mathbf{c})
     - \mathbf{v}_\theta(\mathbf{x}_t,t,\varnothing)\right),
\end{aligned}
\end{equation}
which recovers the standard CFG update in Eq.~\eqref{eq:cfg-feedback}. Thus, CFG is mathematically equivalent to a proportional state-feedback controller (P-control) acting on the semantic alignment error $\mathbf{e}(t)$. The guidance scale $w$, serving as the constant guidance schedule, directly plays the role of the proportional gain.

This state-feedback perspective, decomposing guidance into the guidance schedule $K_t$ and the direction operator $\Pi_t$, provides a structured framework to understand existing CFG advancements. Many follow-up typical CFG variants can be reinterpreted as specific control laws for modulating either the strength via $K_t$ or the direction via $\Pi_t$.

\noindent\textbf{Guidance Schedule}. We next focus on the guidance schedule component $K_t$. Recall that standard CFG applies a \emph{constant} guidance schedule $w$ to the semantic feedback signal.

However, $K_t$ does not need to be fixed. A prominent example of a dynamic guidance schedule is guidance weight scheduling. Recent work~\cite{wang2024analysis} has shown that replacing the constant gain $w$ with a time-varying schedule $w(t)$ leads to substantial improvements in sample quality and semantic consistency. Under our formulation, this corresponds to choosing a {time-dependent} guidance schedule while the direction operator remains identity:
\begin{equation}
     K_t = w(t), \qquad \Pi_t = I,
     \label{eq:ws-parameters}
\end{equation}
with the same semantic error signal $\mathbf{e}(t)$ in Eq.~\eqref{eq:error_definition}. The resulting closed-loop dynamics are shown in Eq.~\eqref{eq:weigt-scheduler}.

This reveals that the weight-scheduler approach is a time-varying proportional feedback controller, also known in control theory as gain-scheduled control. The key difference from standard CFG is that the guidance schedule $K_t$ is no longer fixed.

From a control perspective, the motivation for a dynamic guidance schedule is clear: in early stages of sampling, the state $\mathbf{x}_t$ is dominated by noise, so applying strong correction (a large $K_t$) may amplify noise rather than semantic alignment. A smaller gain $w(t)$ is therefore desirable at high noise levels. As the sample becomes more structured, the feedback signal becomes more semantically meaningful, and the gain can be safely increased.

\noindent\textbf{Direction Operator}. The direction operator $\Pi_t$ can be combined with a more advanced guidance schedule $K_t$. Whereas weight-schedulers~\cite{wang2024analysis} use a scalar $K_t$ and an identity $\Pi_t$, Adaptive Projected Guidance (APG)~\cite{sadat2024APG} demonstrates a case where $K_t$ becomes a matrix gain, working in conjunction with a structured $\Pi_t$. Within the first-order state-feedback framework, APG can be written as:

\begin{equation}
     K_t = w\begin{bmatrix} I\;\; \eta I \end{bmatrix}, \quad \Pi_t = \begin{bmatrix} I - P_t \\ P_t \end{bmatrix}, \quad P_t = \frac{\mathbf{v}_\theta(\mathbf{c})\mathbf{v}_\theta(\mathbf{c})^\top}{|\mathbf{v}_\theta(\mathbf{c})|^2},
     \label{eq:apg-parameters}
\end{equation}
where $P_t$ is an orthogonal projection onto the conditional direction $\mathbf{v}_\theta(\mathbf{c})=\mathbf{v}_\theta(\mathbf{x}_t, t, \mathbf{c})$. Applying the direction operator $\Pi_t$ first decomposes the semantic error $\mathbf{e}(t)$ into orthogonal and parallel components:
\begin{equation}
     \begin{bmatrix}
         \Delta \mathbf{v}_t^{\perp} \\
         \Delta \mathbf{v}_t^{\parallel}
     \end{bmatrix}
     =
     \Pi_t(\mathbf{e}(t))
     =
     \begin{bmatrix}
         I - P_t \\
         P_t
     \end{bmatrix} \mathbf{e}(t).
\end{equation}
The guidance schedule $K_t$, now a structured matrix, applies the global CFG scaling $w$ while introducing an additional factor $\eta$ specifically on the parallel component, yielding:
\begin{equation}
\begin{aligned}
     \frac{d\mathbf{x}_t}{dt}
     &= \mathbf{v}_\theta(\mathbf{x}_t,t,\varnothing) + K_t\,\Pi_t(\mathbf{e}(t)) \\
     &= \mathbf{v}_\theta(\mathbf{x}_t,t,\varnothing)
     + w\left(\Delta \mathbf{v}_t^{\perp} + \eta\,\Delta \mathbf{v}_t^{\parallel}\right).
\end{aligned}
\label{eq:apg-feedback}
\end{equation}

APG therefore reshapes the feedback signal: instead of uniformly amplifying the semantic correction (as in scalar $K_t$), the matrix-based guidance schedule $K_t$ selectively enhances the parallel component aligned with the conditional direction. In control-theoretic terms, APG is a projection-based feedback controller. This design improves semantic alignment without the instability of simply increasing the proportional gain $w$, since it adjusts \emph{how strongly} the guidance acts on different components (via $K_t$) rather than just \emph{how strongly overall}. We show more control interpretations of various CFG methods in Table~\ref{tab:cfg-family}. For notation list and theoretical details, please refer to supplementary material.

\begin{algorithm}[t]
\caption{SMC-CFG}
\label{alg:smc-cfg}
\begin{algorithmic}[1]
\STATE \textbf{Input:} Velocity model $\mathbf{v}_\theta(\cdot ,t,\mathbf{c})$, input condition $\mathbf{c}$, guidance scale $w$, SMC parameters $\lambda$, $k$.
\STATE $\mathbf{x}_T\sim \mathcal{N}(0, \mathbf{I})$
\FOR{$t = T$ \textbf{to} $1$}
    \STATE $\mathbf{v}_t(\mathbf{c}) \leftarrow \mathbf{v}_\theta(\mathbf{x}_t, t, \mathbf{c})$ \hfill \# Conditional prediction
    \STATE $\mathbf{v}_t(\varnothing) \leftarrow \mathbf{v}_\theta(\mathbf{x}_t, t, \varnothing)$ \hfill \# Unconditional prediction
    
    \STATE $\mathbf{e}(t) \leftarrow \mathbf{v}_t(\mathbf{c}) - \mathbf{v}_t(\varnothing)$
    
    \IF{$\mathbf{e}(t+1)$ is None}
        \STATE $\mathbf{e}(t+1) \leftarrow \mathbf{e}(t)$
    \ENDIF
    \STATE $\mathbf{s}_t \leftarrow (\mathbf{e}(t) - \mathbf{e}(t+1)) + \lambda \cdot \mathbf{e}(t+1)$ \hfill \# Sliding surface
    \STATE $\Delta\mathbf{e} \leftarrow -k \cdot \text{sign}(\mathbf{s}_t)$ \hfill \# Switching control
    \STATE $\mathbf{e}(t) \leftarrow \mathbf{e}(t) + \Delta\mathbf{e}$ \hfill \# SMC guidance update

    \STATE $\hat{\mathbf{v}}_t \leftarrow \mathbf{v}_t(\varnothing) + w \cdot \mathbf{e}(t)$
    \STATE $\hat{\mathbf{x}}_{t-1} \leftarrow \text{ODEUpdate}(\mathbf{x}_t, \hat{\mathbf{v}}_t, t)$
\ENDFOR
\STATE \textbf{Return} $\mathbf{x}_0$
\end{algorithmic}
\end{algorithm}

\begin{table*}[t]
\centering
\caption{\textbf{Quantitative evaluation of CFG methods.} Lower ($\downarrow$) FID and higher ($\uparrow$) CLIP, Aesthetic, ImageReward, PickScore, HPSv2, HPSv2.1 and MPS scores indicate better performance. Note that Qwen-Image preserves natural image statistics, yielding the lowest FID.}
\vspace{-3mm}
\resizebox{\linewidth}{!}{
    \begin{tabular}{l|cccccccc}
      \toprule
      \textbf{Guidance}       & \textbf{FID} $\downarrow$ & \textbf{CLIP} $\uparrow$ & \textbf{Aesthetic} $\uparrow$ & \textbf{ImageReward} $\uparrow$ & \textbf{PickScore} $\uparrow$ & \textbf{HPSv2} $\uparrow$ & \textbf{HPSv2.1} $\uparrow$ & \textbf{MPS} $\uparrow$ \\
      \midrule
      
      SD3.5~\cite{esser2024sd3} & 41.725 & 0.3399 & 5.4256 & 0.3591 & 0.2124 & 0.2710 & 0.2372 & 6.5554 \\
      \midrule
      w/ CFG~\cite{ho2022cfg} & 21.421 & 0.3681 & 5.5875 & 0.8889 & 0.2190 & 0.2930 & 0.2842 & 7.2476 \\
      w/ CFG-Zero$^\star$~\cite{fan2025cfg-zero} & 20.317 & 0.3691 & \textbf{5.6124} & 0.9312 & 0.2195 & {0.2942} & {0.2862} & 7.0430 \\
      w/ Rect-CFG++~\cite{saini2025reccfg++} & 20.550 & 0.3655 & 5.5663 & 0.7097 & 0.2173 & 0.2887 & 0.2748 & 6.7854 \\
      {w/ \textbf{SMC-CFG}} & \textbf{20.044} & \textbf{0.3694} & 5.5790 & \textbf{0.9486} & \textbf{0.2211} & \textbf{0.2945 }& \textbf{0.2875} & \textbf{7.5719} \\
      \midrule
      
      Flux-dev~\cite{flux2024} & 52.598 & 0.3272 & 5.4568 & 0.2572 & 0.2137 & 0.2650 & 0.2280 & 5.9592 \\
      \midrule
      w/ CFG~\cite{ho2022cfg} & 27.323 & 0.3692 & 5.5397 & 0.8749 & 0.2228 & 0.2917 & 0.2828 & 7.8387  \\
      w/ CFG-Zero$^\star$~\cite{fan2025cfg-zero} & 26.901 & {0.3742} & 5.7053 & {1.0300} & 0.2262 & \textbf{0.2987} & 0.2992 & 8.1573 \\
      w/ Rect-CFG++~\cite{saini2025reccfg++} & 27.219 & 0.3728 & 5.6909 & 1.0075 & 0.2252 & 0.2974 & 0.2963 & 7.9746 \\
      {w/ \textbf{SMC-CFG}} & \textbf{26.398} & \textbf{0.3743} & \textbf{5.7342} & \textbf{1.0558} & \textbf{0.2268} & 0.2986 & \textbf{0.3021} & \textbf{8.2307} \\
      \midrule
      
      Qwen-Image~\cite{wu2025qwen-image} & 24.894 & 0.3626 & 5.4081 & 0.5742 & 0.2157 & 0.2815 & 0.2613 & 6.7152 \\
      \midrule
      w/ CFG~\cite{ho2022cfg} & 35.431 & 0.3815 & 5.5995 & 1.1063 & 0.2260 & 0.2996 & 0.3038 & 8.1852  \\
      w/ CFG-Zero$^\star$~\cite{fan2025cfg-zero} & 35.391 & 0.3822 & \textbf{5.6598} & 1.1941 & \textbf{0.2279} & 0.3019 & 0.3092 & 8.3739 \\
      w/ Rect-CFG++~\cite{saini2025reccfg++} & 34.371 & 0.3834 & 5.6007 & 1.1727 & 0.2276 & 0.3017 & 0.3068 & 8.1026 \\
      {w/ \textbf{SMC-CFG}} & \textbf{33.371} & \textbf{0.3856} & 5.6289 & \textbf{1.2035} & 0.2275 & \textbf{0.3026} & \textbf{0.3105} & \textbf{8.4320} \\
      \bottomrule
    \end{tabular}
    \vspace{-10mm}
\label{tab:coco_comparison}
}
\end{table*}

\subsection{Sliding Mode Control CFG}
\label{sec:smc-cfg}
Existing CFG methods primarily rely on linear feedback, such as linear combinations or orthogonal projections of the conditional and unconditional velocity estimates. However, the ODE flow is inherently a highly nonlinear dynamical system, particularly when the model capacity becomes large and guidance scale is high. In such regimes, linear guidance tends to amplify nonlinear distortions, leading to oversaturated textures and semantic inconsistency.

To address these issues, we reinterpret CFG under the first-order state-feedback control framework introduced in Sec.~\ref{sec:formulation}. Under this perspective, we propose Sliding Mode Control CFG (SMC-CFG), which introduces a nonlinear sliding surface that continuously corrects the semantic deviation while constraining the diffusion trajectory to evolve toward a stable low-energy semantic manifold.

For the semantic error $\mathbf{e}(t)$ defined in Eq.~\eqref{eq:error_definition}, the ideal target behavior is that $\left (\mathbf{e}(t),\dot{\mathbf{e}}(t)\right)$ decays directly toward origin, as shown in Figure~\ref{fig:convergence} (left):
\begin{equation}
\dot{\mathbf{e}}(t) = -\lambda_0\,\mathbf{e}(t), \quad \lambda > 0.
\label{eq:ideal-exp-decay}
\end{equation}
Here $\lambda_0$ denotes the slope of the ideal line, with its value typically determined by the initial state $(\mathbf{e}(T),\dot{\mathbf{e}}(T))$. The ODE solution $\mathbf{e}(t)=\mathbf{e}(T)\mathrm{exp}(-\lambda t)$ also ensures smooth, monotonic exponential convergence.

However, the diffusion dynamics cannot ensure the ideal process of $\mathbf{e}$; thus we define the sliding mode surface:
\begin{equation}
\mathbf{s}(t) = \dot{\mathbf{e}}(t) + \lambda \mathbf{e}(t),
\label{eq:sliding-surface}
\end{equation}
where $\lambda$ is an adjustable shape parameter of the sliding mode surface, and the surface implicitly encodes the target error dynamics in Eq.~\eqref{eq:ideal-exp-decay}. The manifold $\mathbf{s}(t)=\mathbf{0}$, as illustrated by the dashed line in Figure~\ref{fig:convergence} (right), defines the desired semantic equilibrium flow. We adopt the Lyapunov function~\cite{lyapunov1992general} $V$ to measure the deviation of the system from the sliding manifold. For stable convergence, the system energy must monotonically decrease over time:
\begin{equation}
V(\mathbf{s}) = \tfrac{1}{2}\|\mathbf{s}\|^2, \quad \dot{V} = \mathbf{s}^\top \dot{\mathbf{s}} < 0.
\label{eq:lyapunov}
\end{equation}
We derive $\dot{\mathbf{s}}$ from the semantic guidance error in Eq.~\eqref{eq:error_definition}. Using the chain rule, its time derivative is
\begin{equation}
\begin{split}
\dot{\mathbf{e}}(t)
= \frac{\partial\mathbf{v}_\theta(\mathbf{x},t,\mathbf{c})}{\partial\mathbf{x}}\dot{\mathbf{x}}
- \frac{\partial\mathbf{v}_\theta(\mathbf{x},t,\varnothing)}{\partial\mathbf{x}}\dot{\mathbf{x}} \\
+ \frac{\partial\mathbf{v}_\theta(\mathbf{x},t,\mathbf{c})}{\partial t}
- \frac{\partial\mathbf{v}_\theta(\mathbf{x},t,\varnothing)}{\partial t}.
\end{split}
\end{equation}
We introduce a sliding-mode correction term $\Delta \mathbf{e}(t)$, giving the full control $\mathbf{u}(t)=w(\mathbf{e}(t)+\Delta \mathbf{e}(t))$, which modulates the semantic guidance by directly shaping the error dynamics rather than altering the model prediction.

Substituting the controlled state dynamics in Eq.~\eqref{eq:general-feedback} into the time derivative of the semantic error yields:
\begin{equation}
\resizebox{0.9\hsize}{!}{$
\dot{\mathbf{e}}(t)=\mathbf{\Phi}(t,\mathbf{x})\\
+w\Big(\frac{\partial\mathbf{v}_\theta(\mathbf{x},t,\mathbf{c})}{\partial\mathbf{x}}
-\frac{\partial\mathbf{v}_\theta(\mathbf{x},t,\varnothing)}{\partial\mathbf{x}}\Big)\mathbf{G}(\mathbf{x},t)\,\Delta\mathbf{e}(t),$}
\label{eq:error_diff}
\end{equation}
where $\mathbf{\Phi}$ absorbs terms independent of $\Delta\mathbf{e}(t)$. Differentiating and substituting the sliding surface definition in Eq.~\eqref{eq:sliding-surface}, we obtain:
\begin{equation}
\dot{\mathbf{s}}(t)=\mathbf{\Phi}_s(t,\mathbf{x})+\mathbf{\Gamma}_s(t)\,\Delta \mathbf{e}(t),
\end{equation}
where $\mathbf{\Gamma}_s$ denotes the coefficient matrix multiplying $\Delta \mathbf{e}$, and $\mathbf{\Phi}_s(t,\mathbf{x})$ represents all remaining terms. We assume that $\mathbf{\Gamma}_s$ has minimum singular value lower-bounded (\ie, $\sigma_{\min}(\mathbf{\Gamma}_s)\ge b_{\min}>0$), and $\mathbf{\Phi}_s$ is bounded (\ie, $\|\mathbf{\Phi}_s\|\le\delta$, $\exists \delta >0$), which are standard in sliding mode control.

Substituting into the Lyapunov derivative yields:
\begin{equation}
\dot{V} = \mathbf{s}^\top \mathbf{\Phi}(t,\mathbf{e}) + \mathbf{s}^\top \mathbf{\Gamma}(t)\Delta \mathbf{e}(t).
\end{equation}

We apply the classical switching control law:
\begin{equation}
\Delta \mathbf{e}(t) = -\mathbf{K}\cdot\mathrm{sign}(\mathbf{s}(t)),
\label{eq:smc-law}
\end{equation}
where $\mathbf{K} = k\,\mathbf{I}$ is a positive diagonal gain matrix. Since $\sigma_{\min}(\mathbf{\Gamma}(t)\mathbf{K}) \ge k\,b_{\min}$, we obtain:
\begin{equation}
\dot{V}\le \|\mathbf{s}\|\delta - k\,b_{\min}\|\mathbf{s}\|
= -(k\,b_{\min}-\delta)\|\mathbf{s}\|.
\end{equation}
Therefore, choosing $k$ such that $k\,b_{\min}>\delta$ ensures:
\begin{equation}
\dot{V} =\mathbf{s}^\top \dot{\mathbf{s}} \le -\eta\|\mathbf{s}\|,\qquad \eta = k\,b_{\min}-\delta > 0.
\end{equation}
Dividing both sides by \(\|\mathbf{s}\|>0\) yields the scalar differential inequality
\[
\frac{d}{dt}\|\mathbf{s}(t)\| \le -\eta.
\]
Integrating from $0$ to $t$ gives
\[
\|\mathbf{s}(t)\| \le \|\mathbf{s}(0)\| - \eta t,
\]
which supports finite-time convergence of $\mathbf{s}(t)$ to zero. In particular:
\begin{equation}
\|\mathbf{s}(t)\| = 0 \quad \text{for some} \quad t \le \frac{\|\mathbf{s}(0)\|}{\eta}.
\end{equation}

We present the entire method in Algorithm~\ref{alg:smc-cfg}. The proposed sliding mode surface and switching control law enforce stable semantic guidance by ensuring that the diffusion trajectory converges to the desired manifold, depicted by the red curve in Figure~\ref{fig:convergence}, eliminating oscillations and improving consistency during guided sampling.

\section{Experiments}
\subsection{Experimental Setups}

\begin{figure*}[t!]
    \centering
    \includegraphics[width=\textwidth]{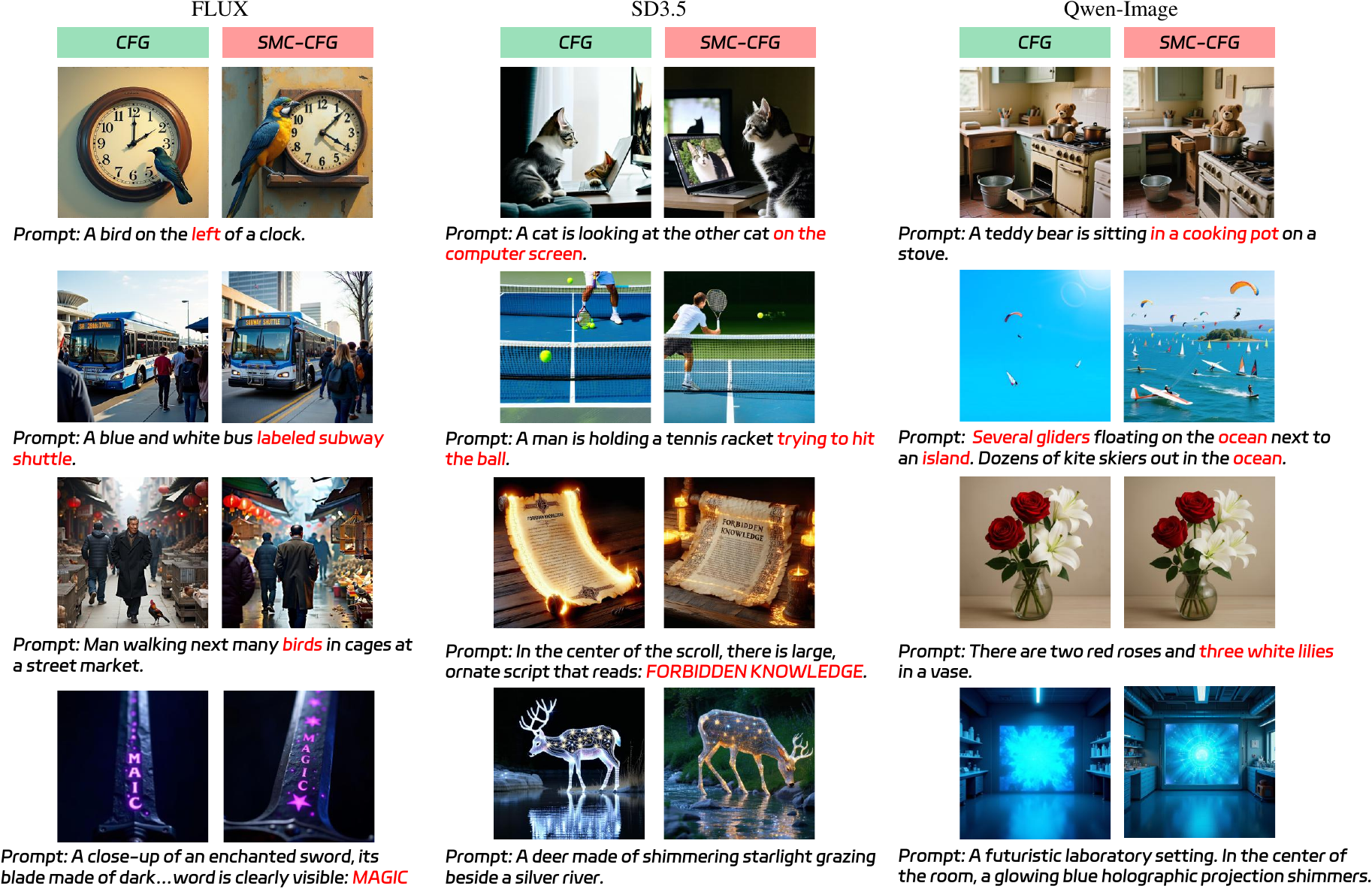}
    \vspace{-5mm}
    \caption{\textbf{Qualitative results across different T2I models.} We provide visual comparisons between CFG and our SMC-CFG across various models. SMC-CFG exhibits better performance in positional relationships, text generation, and detailed object representation.}
    \vspace{-5mm}
    \label{fig:comparison_among_models}
\end{figure*}

\noindent\textbf{Datasets and Baselines.}
We conduct experiments on a subset of the MS-COCO~\cite{lin2014mscoco} dataset, comprising 5,000 image-text pairs. To demonstrate the generality of our method across diverse model scales, we evaluate it on several state-of-the-art flow-based T2I models, including Stable Diffusion 3.5 (SD3.5)~\cite{esser2024sd3}, Flux-dev~\cite{flux2024}, and Qwen-Image~\cite{wu2025qwen-image} with 8B, 12B, and 20B parameters, respectively. In addition to comparing against the standard Classifier-Free Guidance baseline, we include two recent guidance variants designed specifically for flow-matching generative models: CFG-zero$^\star$~\cite{fan2025cfg-zero} and Rectified-CFG++~\cite{saini2025reccfg++}. We implement both methods on all evaluated backbones following the official paper and open-source code to ensure a fair and consistent comparison. For more comprehensive experiments on additional T2I benchmark and diffusion model, please refer to our supplementary material.

\noindent\textbf{Evaluation Metrics.}
To assess image quality and visual realism, we report the Fr\'{e}chet Inception Distance (FID)~\cite{heusel2017FID}. To measure the alignment between generated images and input text prompts, we use the CLIP Score~\cite{radford2021clip,hessel2021clipscore}, which quantifies semantic consistency in the joint vision–language embedding space. In addition to these core metrics, we further provide a comprehensive evaluation of aesthetic quality and human preference, including Aesthetic Score~\cite{schuhmann2022laionAesthetics}, ImageReward~\cite{xu2023imagereward}, PickScore~\cite{kirstain2023pickscore}, HPSv2~\cite{wu2023HPSv2}, HPSv2.1~\cite{wu2023HPSv2}, and MPS~\cite{zhang2024MPS}. Together, these metrics offer a holistic perspective on both the fidelity and human-perceived appeal of the generated content.

\noindent\textbf{Implementation Details.}
All experiments are conducted on a single NVIDIA A100 GPU (40GB). We implement the proposed method SMC-CFG on three representative pretrained T2I diffusion models: SD3.5~\cite{esser2024sd3}, Flux-dev~\cite{flux2024}, and Qwen-Image~\cite{wu2025qwen-image}. For all models, we adopt their default CFG scales provided in the official implementations. In our SMC-CFG framework, the two hyperparameters $\lambda$ and $K$ are kept fixed within each model and shared across all datasets and experimental conditions to ensure fair comparison. See supplementary material for more implementation details and complete hyperparameter configurations.

\subsection{Text-to-Image Generation}
In this section, we evaluate the effectiveness of our proposed SMC-CFG in text-to-image generation. Experiments are conducted on the MS-COCO~\cite{lin2014mscoco} dataset using three state-of-the-art flow-based models. To ensure a fair and up-to-date comparison, we implement two recent CFG variants (CFG-zero$^\star$~\cite{fan2025cfg-zero} and Rectified-CFG++~\cite{saini2025reccfg++}) designed for flow-matching models as baselines.

\noindent\textbf{Quantitative Evaluation.}
Table~\ref{tab:coco_comparison} reports the quantitative results of SMC-CFG compared with the standard CFG and baselines across different T2I models. Our method consistently achieves lower FID scores, indicating the generated images exhibit improved visual quality and realism. Meanwhile, the higher CLIP Scores demonstrate stronger semantic alignment between the generated images and the input prompts. Furthermore, SMC-CFG attains superior scores on ImageReward, HPSv2.1, and MPS scores, signifying that the generated images are more aligned with human aesthetic and preference judgments. For additional metrics, our method also achieves comparable or better results than the baselines, demonstrating strong overall generation quality.

\begin{figure}
    \centering
    \includegraphics[width=\linewidth]{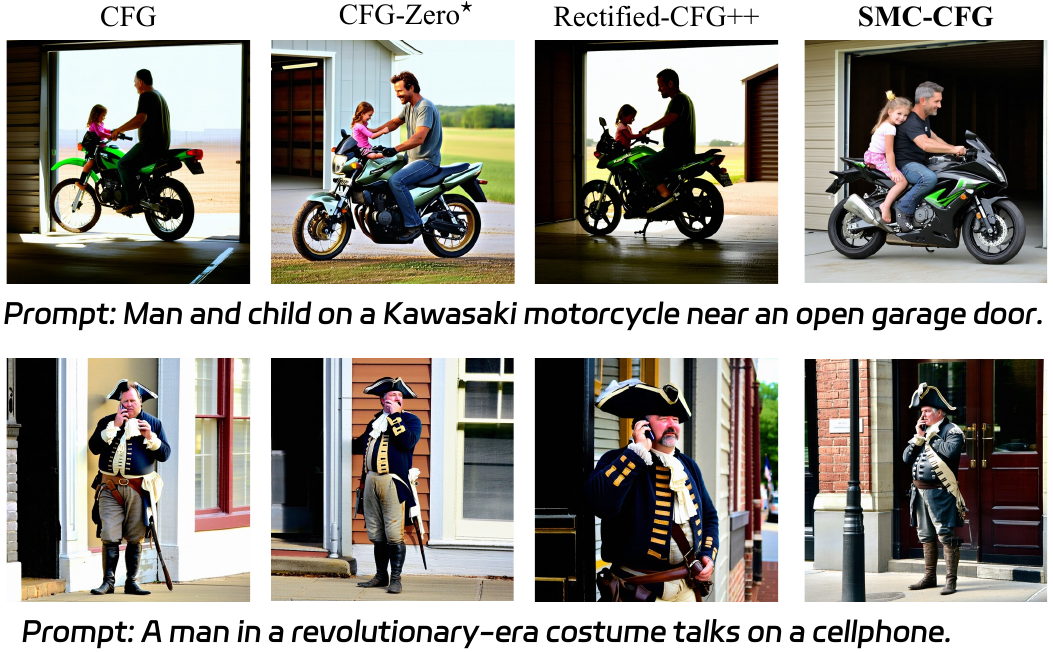}
    \vspace{-6mm}
    \caption{\textbf{Qualitative comparison with baseline methods.} For challenging scenarios including relative positions, clothing styles, and human actions, baseline methods produce irrational outputs, while SMC-CFG preserves robust text consistency.}
    \vspace{-6mm}\label{fig:compare_with_baselines}
\end{figure}

\noindent\textbf{Qualitative Evaluation.}
We further present qualitative comparisons to illustrate the improvements achieved by SMC-CFG. As shown in Figure~\ref{fig:comparison_among_models}, across different model backbones, our method produces images with sharper details, more coherent object structures, and more faithful adherence to the textual descriptions compared to standard CFG. This demonstrates that our approach is consistently effective and model-agnostic. In addition, Figure~\ref{fig:compare_with_baselines} highlights results on more challenging prompts involving complex compositions, fine-grained semantics, or stylistic attributes. Compared with recent flow-matching-based guidance variants, SMC-CFG generates images that better preserve semantic correctness and maintain aesthetic quality, without introducing over-smoothing or mode collapse.

\begin{table}[htbp]
\centering
\caption{\textbf{Ablation study on hyperparameter $\lambda$ and $k$.} We conduct ablation across various hyperparameter settings in four metrics: FID, CLIP, Aesthetic (Aesth), and ImageReward (ImgRwd), respectively measuring generation quality, semantic alignment, aesthetic level, and human preference.}
\vspace{-2mm}
\label{tab:ablation}
\small
\setlength{\tabcolsep}{8pt}  
\renewcommand{\arraystretch}{1.05}  
\begin{tabular}{ll|cccc}
\toprule
\textbf{$\lambda$} & \textbf{$k$} & \textbf{FID} $\downarrow$ & \textbf{CLIP} $\uparrow$ & \textbf{Aesth} $\uparrow$ & \textbf{ImgRwd} $\uparrow$ \\
\midrule
3 & 0.1 & 26.193 & 0.3698 & 5.7064 & 1.0174 \\
4 & 0.1 & 26.006 & 0.3701 & 5.7098 & 1.0219 \\
5 & 0.1 & \textbf{25.951} & \textbf{0.3709} & \textbf{5.7128} & \textbf{1.0248} \\
6 & 0.1 & 26.143 & 0.3703 & 5.7071 & 1.0228 \\
\midrule
5 & 0.1 & \textbf{25.951} & 0.3709 & 5.7128 & 1.0248 \\
5 & 0.4 & 26.143 & 0.3719 & \textbf{5.7218} & \textbf{1.0504} \\
5 & 0.7 & 26.416 & 0.3739 & 5.7175 & 1.0406 \\
5 & 1.0 & 26.281 & \textbf{0.3741} & 5.7054 & 1.0453 \\
\bottomrule
\end{tabular}
\vspace{-3mm}
\end{table}

\subsection{Ablation Studies and Analysis}

\noindent\textbf{Ablation on Hyperparameters.}
To gain a deeper understanding of the roles of hyperparameters in SMC-CFG, we perform an ablation study on their distinct impacts. The top of Table~\ref{tab:ablation} illustrates how $\lambda$ shapes the sliding mode surface. Extreme values (too low or too high) distort this manifold, impairing guidance stability and diminishing output fidelity. On the bottom of Table~\ref{tab:ablation}, we explore influence of $k$ (with fixed $\lambda$), which governs the force toward the sliding mode surface. Modest $k$ yields slow convergence and meanwhile alleviates the distortions introduced by CFG, thereby weakening text-image alignment (\ie, lower CLIP scores) but preserves realism (\ie, lower FID). In contrast, excessive $k$ causes abrupt pulls, triggering erratic sampling or vibrations. Though boosting semantic match, such outputs suffer from reduced aesthetic appeal and poor human-preference ratings. Overall, suitable hyperparameters strike a trade-off between perceptual excellence and textual fidelity.

\begin{figure}[t]
    \centering
    \includegraphics[width=\linewidth]{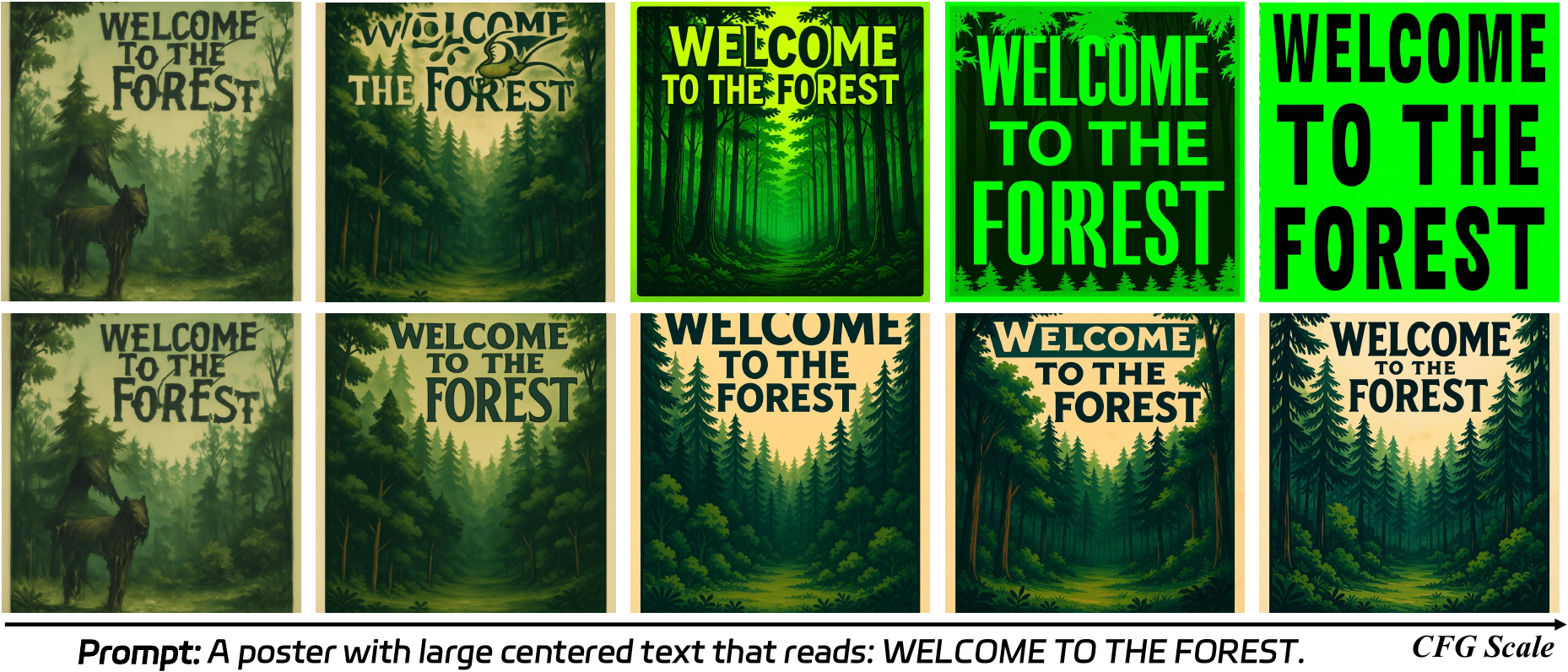}
    \vspace{-6.5mm}
    \caption{Visual comparison between CFG (top) and SMC-CFG (bottom) across different CFG scales.}
    \vspace{-6.3mm}
    \label{fig:cfg_scale}
\end{figure}

\noindent\textbf{Guidance Scale.}
We analyze how guidance scales influence the generation performance of SMC-CFG in Figure~\ref{fig:cfg_scale}. At large guidance scales, CFG improves semantic alignment at the cost of significant degradation in image quality and realism. In contrast, SMC-CFG exhibits more stable performance across a wide range of guidance scales, maximizing the capability of CFG while avoiding significant reductions in image quality and aesthetic appeal.

\section{Conclusion}
We explore a unified framework called CFG-Ctrl, reinterpreting CFG as a feedback control in flow matching models and analyzing its nonlinear behaviors under high guidance scales. From this perspective, we further propose SMC-CFG, a nonlinear control-based guidance mechanism that introduces a switching control term to enforce fast and stable convergence along the sliding mode surface. Extensive experiments demonstrate that SMC-CFG consistently improves semantic alignment and visual fidelity while maintaining robustness across diverse guidance scales. Ablation studies also reveal how its hyperparameters affect stability and perception. We believe this control-theoretic perspective provides a promising direction for more effective and robust guidance in future large-scale generative models.
{
    \small
    \bibliographystyle{ieeenat_fullname}
    \bibliography{main}
}

\clearpage
\setcounter{page}{1}
\maketitlesupplementary

\section{More Theoretical Analysis}
\subsection{Notation Table}
To facilitate the understanding of the theoretical derivations of CFG-Ctrl and SMC-CFG, we summarize the main symbols, their corresponding technical meanings, and relevant mathematical expressions or value constraints in Table~\ref{tab:notation_table}. These notations cover core components such as velocity fields, semantic error signals, and stability analysis metrics, providing a clear reference for readers to follow the logical flow of the proposed framework.

\begin{table*}[ht]
\centering
\caption{\textbf{Notation table.}} \label{tab:notation_table}
\vspace{-1.5mm}
\resizebox{\linewidth}{!}{
\begin{tabularx}{\textwidth}{lXl}
    \toprule Notation & Meaning & Value \\
    \midrule 
    $\mathbf{x}_t$  & Latent state at time $t$ during generative flow sampling. & $\mathbf{x}_0 \sim \mathcal{N}(0, \mathbf{I})$ (initial state) \\
    \midrule
    $\mathbf{v}_{\theta}(\mathbf{x}_t, t, \varnothing)$  & Unconditional velocity field, obtained by dropping the condition $\mathbf{c}$. & / \\
    \midrule
    $\mathbf{v}_{\theta}(\mathbf{x}_t, t, \mathbf{c})$  & Conditional velocity field, incorporating the input condition $\mathbf{c}$. & / \\
    \midrule
    $\hat{\mathbf{v}}_{\theta}(\mathbf{x}_t, t, \mathbf{c})$  & Guided velocity field, combined via guidance. & / \\
    \midrule
    $w$  & CFG guidance scale. & $w \geq 1$ \\
    \midrule
    $\mathbf{e}(t)$  & Semantic error signal. & $\mathbf{v}_{\theta}(\mathbf{x}_t, t, \mathbf{c}) - \mathbf{v}_{\theta}(\mathbf{x}_t, t, \varnothing)$ \\
    \midrule
    $\dot{\mathbf{e}}(t)$  & Temporal derivative of the semantic error signal. & / \\
    \midrule
    $\mathbf{u}_t$  & General guidance control input. & $\mathbf{u}_t = K_t \Pi_t(\mathbf{e}(t))$ \\
    \midrule
    $K_t$  & Guidance schedule matrix/scalar in CFG-Ctrl framework. & / \\
    \midrule
    $\Pi_t$  & Direction operator in CFG-Ctrl framework. & / \\
    \midrule
    $\mathcal{S}$ & Semantic sliding manifold. & $\mathcal{S} = \{ (\mathbf{x}, t) \mid \mathbf{s}(t) = \mathbf{0} \}$ \\
    \midrule
    $\mathbf{s}(t)$  & Sliding mode surface variable in SMC-CFG. & $\mathbf{s}(t) = \dot{\mathbf{e}}(t) + \lambda \mathbf{e}(t)$ \\
    \midrule
    $\lambda$  & Shape parameter of the sliding mode surface. & Hyperparameter \\
    \midrule
    $k$  & Gain of the switching control term. & Hyperparameter \\
    \midrule
    $\Delta \mathbf{e}(t)$  & SMC correction term (Switching Control). & $- k \cdot \text{sign}(\mathbf{s}(t))$ \\
    \midrule
    $\mathbf{\Phi}(t, \mathbf{x}_t)$ & Intrinsic drift dynamics (encapsulating model non-linearities). & $\|\mathbf{\Phi}\|_2 \leq \delta$ \\
    \midrule
    $\mathbf{\Gamma}(t)$ & Effective control gain matrix (Jacobian of semantic difference). & $\mathbf{\Gamma} = w \mathbf{I} + \Delta \mathbf{\Gamma}(t)$ \\
    \midrule
    $\Delta \mathbf{\Gamma}(t)$ & Anisotropic deviation from the nominal isotropic guidance. & / \\
    \midrule
    $\delta$ & Upper bound of the intrinsic drift dynamics. & $\delta > 0$ \\
    \midrule
    $\rho$ & Upper bound of the anisotropic deviation norm. & $\|\Delta \mathbf{\Gamma}\|_2 \leq \rho < w$ \\
    \midrule
    $\epsilon$ & Positive safety margin for the control gain. & $\epsilon > 0$ \\
    \midrule
    $V(\mathbf{s})$ & Lyapunov function candidate for stability analysis. & $V(\mathbf{s}) = \frac{1}{2}\|\mathbf{s}\|_2^2$ \\
    \midrule
    $\Delta t$ & Discrete time step size for sampling. & / \\
\hline
\end{tabularx}}
\end{table*}

\subsection{Additional CFG Variants}

\noindent\textbf{CFG-Zero$^\star$~\cite{fan2025cfg-zero}} introduces an optimizable scalar $s^\star \in \mathbb{R}_{>0}$ into the standard CFG framework, with its guided velocity field formulated as:
\begin{equation}
\begin{aligned}
    \label{eq:cfg-zero}
    \hat{\mathbf{v}}_\theta(\mathbf{x}_t, t, \mathbf{c}) &=(1-w)\cdot s^\star \cdot\mathbf{v}_\theta(\mathbf{x}_t, t, \varnothing) + w\,\mathbf{v}_\theta(\mathbf{x}_t, t, \mathbf{c}) \\ 
    &= s^\star \cdot\mathbf{v}_\theta(\mathbf{x}_t, t, \varnothing) \\ & \quad+ w \cdot \bigl( \mathbf{v}_\theta(\mathbf{x}_t, t, \mathbf{c}) - s^\star \cdot\mathbf{v}_\theta(\mathbf{x}_t, t, \varnothing) \bigr). 
\end{aligned}
\end{equation}
As summarized in Table 1 of the main text, under the CFG-Ctrl paradigm, the guidance schedule $K_t$ and direction operator $\Pi_t$ of CFG-Zero$^\star$ are modeled as:
\begin{equation}
\begin{aligned}
    &K_t = \begin{bmatrix} wI & \frac{s^\star}{1-s^\star}I \end{bmatrix},\quad s^\star=\frac{\mathbf{v}_\theta(\mathbf{c})^\top\mathbf{v}_\theta(\mathbf{\varnothing})}{|\mathbf{v}_\theta(\mathbf{\varnothing})|^2}), \\ &\Pi_t = \begin{bmatrix} I - P_t \\ P_t \end{bmatrix},\quad P_t = \frac{\mathbf{v}_\theta(\mathbf{\varnothing})\mathbf{v}_\theta(\mathbf{\varnothing})^\top}{|\mathbf{v}_\theta(\mathbf{\varnothing})|^2}.
\end{aligned}
\end{equation}
Substituting these components into the closed-loop dynamics of CFG-Ctrl yields:
\begin{equation}
\label{eq:cfg-zero-control}
\begin{aligned}
    \frac{d\mathbf{x}_t}{dt}
    &= \mathbf{v}_\theta(\mathbf{x}_t,t, \varnothing) + K_t\,\Pi_t(e_t) \\
    &= \frac{\mathbf{v}^\top_\theta(\mathbf{c})\mathbf{v}_\theta(\mathbf{\varnothing})}{|\mathbf{v}_\theta(\mathbf{\varnothing})|^2}\mathbf{v}_\theta(\mathbf{x}_t,t, \varnothing) \\ & \quad + w\left(\mathbf{v}_\theta(\mathbf{x}_t,t,\mathbf{c}) - \frac{\mathbf{v}^\top_\theta(\mathbf{c})\mathbf{v}_\theta(\mathbf{\varnothing})}{|\mathbf{v}_\theta(\mathbf{\varnothing})|^2}\mathbf{v}_\theta(\mathbf{x}_t,t,\varnothing)\right) \\
    &= s^\star_t \cdot \mathbf{v}_\theta(\mathbf{x}_t,t, \varnothing) + w\left(\mathbf{v}_\theta(\mathbf{x}_t,t,\mathbf{c}) - s^\star_t \cdot \mathbf{v}_\theta(\mathbf{x}_t,t,\varnothing)\right),
\end{aligned}
\end{equation}
where $s^\star_t$ corresponds to $\frac{\mathbf{v}^\top_\theta(\mathbf{c})\mathbf{v}_\theta(\mathbf{\varnothing})}{|\mathbf{v}_\theta(\mathbf{\varnothing})|^2}$. Notably, CFG-Zero$^\star$ shares a similar design motivation with APG~\cite{sadat2024APG}: both adopt orthogonal projection transformations as their direction operators. The key distinction lies in the projection target—APG projects onto the conditional velocity field $\mathbf{v}_\theta(\mathbf{x}_t,t,\mathbf{c})$, while CFG-Zero$^\star$ projects onto the unconditional velocity field $\mathbf{v}_\theta(\mathbf{x}_t,t,\varnothing)$. From a control-theoretic perspective, both methods fall into the category of projection-based structured feedback controllers.

\noindent\textbf{Rectified-CFG++~\cite{saini2025reccfg++}} differs from standard CFG by incorporating not only the error signal derived from the current latent state $\mathbf{x}_t$ (defined as $\Delta \mathbf{v}_\theta(t) = \mathbf{v}_\theta(\mathbf{x}_t, t, \mathbf{c}) - \mathbf{v}_\theta(\mathbf{x}_t, t, \varnothing)$) but also predictive information from a future state $\mathbf{x}_{t-\frac{\Delta t}{2}}$. The error signal for this predicted future state is formulated as:
\begin{equation}
    \Delta \mathbf{v}_\theta(t-\frac{\Delta t}{2}) =  \mathbf{v}_\theta(\mathbf{x}_{t-\frac{\Delta t}{2}}, t-\frac{\Delta t}{2}, \mathbf{c})- \mathbf{v}_\theta(\mathbf{x}_{t-\frac{\Delta t}{2}}, t-\frac{\Delta t}{2}, \varnothing),
\end{equation}
and the guided velocity field of Rectified-CFG++ is given by:
\begin{equation}
    \label{eq:rec-cfg++}
    \hat{\mathbf{v}}_\theta(\mathbf{x}_t, t, \mathbf{c}) = \mathbf{v}_\theta(\mathbf{x}_t, t, \mathbf{c}) + \alpha(t)\,\Delta \mathbf{v}_\theta(t-\frac{\Delta t}{2}).
\end{equation}
As outlined in Table 1 of the main text, within the CFG-Ctrl framework, the guidance schedule $K_t$ and direction operator $\Pi_t$ of Rectified-CFG++ are structured as:
\begin{equation}
\begin{aligned}
    K_t = &\begin{bmatrix}
            I \;\; \alpha(t)I
        \end{bmatrix}, \quad\alpha(t)=\lambda_{max}(1-t)^\gamma,\\ &\quad \Pi_t = \begin{bmatrix}
            \Delta \mathbf{v}_\theta(t) \\
            \Delta \mathbf{v}_\theta(t-\frac{\Delta t}{2})
        \end{bmatrix}.
\end{aligned}
\end{equation}
Substituting these components into the closed-loop dynamics of CFG-Ctrl leads to the following derivation:
\begin{equation}
\label{eq:rect-cfg++-control}
\begin{aligned}
    \frac{d\mathbf{x}_t}{dt}
    &= \mathbf{v}_\theta(\mathbf{x}_t,t, \varnothing) + K_t\,\Pi_t(e_t) \\
    &= \mathbf{v}_\theta(\mathbf{x}_t,t, \varnothing) + \Delta \mathbf{v}_\theta(t) + \alpha(t)\Delta \mathbf{v}_\theta(t-\frac{\Delta t}{2}) \\
    &= \mathbf{v}_\theta(\mathbf{x}_t,t, \mathbf{c}) + \alpha(t)\Delta \mathbf{v}_\theta(t-\frac{\Delta t}{2}).
\end{aligned}
\end{equation}

Notably, Rectified-CFG++ adopts a time-varying gain scheduling strategy via $\alpha(t)$
, which dynamically adjusts guidance strength throughout the sampling process. Beyond this, the method embodies the core principle of Model Predictive Control, which is a robust control paradigm that leverages a system model to predict future behavior over a finite horizon and optimize control actions accordingly. By integrating error information from the predicted future state $\mathbf{x}_{t-\frac{\Delta t}{2}}$, Rectified-CFG++ effectively anticipates potential deviations in the generative flow and pre-emptively adjusts guidance, thereby enhancing the stability of semantic alignment and the efficiency of the sampling process.

\subsection{Theoretical Motivation: Robustness Analysis}
\label{sec:lyapunov_analysis}

In this section, we provide a theoretical motivation for the proposed SMC-CFG framework from a robust control perspective.
Unlike standard CFG, which relies on linear extrapolation and assumes an ideal linear evolution of the semantic error, SMC-CFG explicitly introduces a nonlinear switching term to handle the unmodeled non-linearities and disturbances inherent in the diffusion flow.
Our analysis demonstrates that under reasonable robustness assumptions, the proposed controller drives the generative trajectory toward the \textit{semantic sliding manifold} $\mathcal{S} = \{ (\mathbf{x}, t) \mid \mathbf{s}(t) = \mathbf{0} \}$.

\subsubsection{Dynamics of the Sliding Variable}

Let $\mathbf{e}(t) = \mathbf{v}_\theta(\mathbf{x}_t, t, \mathbf{c}) - \mathbf{v}_\theta(\mathbf{x}_t, t, \varnothing)$ denote the semantic error signal.
Recall that the sliding variable is defined as $\mathbf{s}(t) = \dot{\mathbf{e}}(t) + \lambda \mathbf{e}(t)$.
Substituting the closed-loop update law into the time derivative of the sliding variable, we obtain the governing equation:
\begin{equation}
    \dot{\mathbf{s}}(t) = \mathbf{\Phi}(t, \mathbf{x}_t) + \mathbf{\Gamma}(t) \cdot \Delta \mathbf{e}(t),
    \label{eq:s_dynamics_formal}
\end{equation}
where:
\begin{itemize}
    \item $\mathbf{\Phi}(t, \mathbf{x}_t)$ represents the \textit{intrinsic drift dynamics}, encapsulating the system's natural evolution and standard CFG terms.
    \item $\mathbf{\Gamma}(t)$ denotes the \textit{effective control gain matrix}, which corresponds to the scaled Jacobian of the semantic difference: $\mathbf{\Gamma}(t) = w \nabla_{\mathbf{x}} (\mathbf{v}_{\theta}(\mathbf{c}) - \mathbf{v}_{\theta}(\varnothing))$.
\end{itemize}

A key challenge in diffusion models is that $\mathbf{\Gamma}(t)$ is highly non-linear and anisotropic. To address this, we adopt a robust control strategy by decomposing the gain into a nominal part and a deviation part.

\subsubsection{Robustness Assumptions}

\begin{assumption}[Boundedness of Intrinsic Drift]
\label{assump:boundedness}
While the gradients of diffusion models may diverge at time boundaries ($t \to 0$ or $t \to T$), we assume that within the effective sampling interval, the drift term $\mathbf{\Phi}(t, \mathbf{x}_t)$ is locally bounded:
\begin{equation}
    \sup_{t, \mathbf{x} \in \mathcal{D}} \| \mathbf{\Phi}(t, \mathbf{x}) \|_2 \le \delta.
    \label{eq:boundedness_condition}
\end{equation}
\end{assumption}

\begin{assumption}[Nominal Control Dominance]
\label{assump:dominance}
We decompose the effective gain matrix $\mathbf{\Gamma}(t)$ into a nominal isotropic gain $w\mathbf{I}$ and an anisotropic deviation $\Delta \mathbf{\Gamma}(t)$:
\begin{equation}
    \mathbf{\Gamma}(t) = w \mathbf{I} + \Delta \mathbf{\Gamma}(t).
\end{equation}
We assume that the guidance scale $w$ is sufficiently large such that the nominal control direction dominates the anisotropic deviation, in the sense that there exists a constant $\rho > 0$ with $w > \rho \sqrt{D}$ where the constant $D$ is the dimension of $\mathbf{s}$ . And the spectral norm of the deviation is bounded:
\begin{equation}
    \| \Delta \mathbf{\Gamma}(t) \|_2 \le \rho.
    \label{eq:dominance_condition}
\end{equation}
\end{assumption}
\noindent\textit{Remark:} Assumption \ref{assump:dominance} is physically intuitive: it implies that the CFG guidance force $w$ remains the dominant driver of the semantic correction, while the local curvature of the velocity field $\Delta \mathbf{\Gamma}$ acts as a subordinate disturbance.

\subsubsection{Robust Stability Analysis}

We now show that the proposed switching control law $\Delta \mathbf{e}(t) = -k \cdot \text{sign}(\mathbf{s}(t))$ ensures stability despite these uncertainties.

\begin{theorem}[Robust Convergence]
\label{thm:finite_time}
Consider the system in Eq.~\eqref{eq:s_dynamics_formal} under Assumptions \ref{assump:boundedness} and \ref{assump:dominance}. If the switching gain $k$ satisfies:
\begin{equation}
    k > \frac{\delta}{w - \rho \sqrt{D}} + \epsilon,
    \label{eq:gain_condition}
\end{equation}
where $\epsilon > 0$ is a safety margin.
\end{theorem}

Consider the Lyapunov function $V(\mathbf{s}) = \frac{1}{2} \|\mathbf{s}\|_2^2$. Its derivative is:
\begin{equation}
    \dot{V} = \mathbf{s}^\top \dot{\mathbf{s}} = \mathbf{s}^\top \left( \mathbf{\Phi} + (w\mathbf{I} + \Delta \mathbf{\Gamma}) \Delta \mathbf{e} \right).
\end{equation}
Substituting the control law $\Delta \mathbf{e} = -k \cdot \text{sign}(\mathbf{s}(t))$ (for $\mathbf{s} \neq \mathbf{0}$):
\begin{equation}
\begin{aligned}
    \dot{V} &= \mathbf{s}^\top \mathbf{\Phi} - w k \|\mathbf{s}\|_1 - k \mathbf{s}^\top \Delta \mathbf{\Gamma} \cdot \text{sign}(\mathbf{s}(t)) \\
            &\le \|\mathbf{s}\|_2 \|\mathbf{\Phi}\|_2 - w k \|\mathbf{s}\|_1 + k \|\mathbf{s}\|_2 \|\text{sign}(\mathbf{s}(t))\|_2 \|\Delta \mathbf{\Gamma}\|_2 \\
            &\le \delta \|\mathbf{s}\|_2 - w k \|\mathbf{s}\|_1 + k \rho \sqrt{D}\,\|\mathbf{s}\|_2,
\end{aligned}
\end{equation}
Let $\phi = \omega - \rho \sqrt{D}$ and apply the bounds $\delta$ and $\rho$ from Assumptions \ref{assump:boundedness} and \ref{assump:dominance}:
\begin{equation}
    \dot{V} \le \|\mathbf{s}\|_2 \left( \delta - w k + k \rho \sqrt{D} \right) = \|\mathbf{s}\|_2 \left( \delta - k \phi \right).
\end{equation}
From the condition in Eq.~\eqref{eq:gain_condition}, we have $k \phi > \delta + \epsilon \phi$. Substituting this into the inequality:
\begin{equation}
    \dot{V} \le \|\mathbf{s}\|_2 \left( \delta - (\delta + \epsilon \phi) \right) = - \epsilon \phi \|\mathbf{s}\|_2.
\end{equation}
Let $\eta = \epsilon \phi > 0$. The differential inequality $\dot{V} \le -\sqrt{2}\eta V^{1/2}$ guarantees finite-time convergence of $\mathbf{s}(t) $.

This analysis demonstrates that SMC-CFG is theoretically robust: as long as the gain $k$ is chosen to cover the worst-case combination of intrinsic drift $\delta$ and the dimension-amplified Jacobian mismatch $\rho \sqrt{D}$ induced by the sign-based switching law, the system remains stable.

\subsubsection{Discrete Implementation and Stability Corridor}

The theoretical derivation above serves as a continuous-time design guide. In practice, diffusion models operate in discrete time steps $\Delta t$, where high-gain switching can lead to chattering.
Based on the discrete evolution $\|\mathbf{s}_{t+1}\| \approx | \|\mathbf{s}_t\| - \Delta t (w_{eff} k - \delta) |$, we derive a heuristic \textbf{Stability Corridor} for hyperparameter tuning:
\begin{equation}
    \underbrace{\frac{\delta_{est}}{w}}_{\text{Convergence}} < k < \underbrace{\frac{2 \|\mathbf{s}_t\|_2}{w \Delta t}}_{\text{Stability}}.
    \label{eq:stability_corridor}
\end{equation}
This corridor highlights the trade-off: $k$ must be large enough to overcome model drift (lower bound), but bounded by the inverse step size to prevent numerical oscillations (upper bound). This aligns with our experimental findings in Table 3, where a moderate fixed $k$ achieves the optimal balance.

\noindent\textbf{Remark on Hyperparameter Selection.} The theoretical analysis in Eq.~\eqref{eq:stability_corridor} establishes a stability corridor for the gain $k$, bounded by the intrinsic model drift $\delta$ (lower bound) and the discretization frequency $1/\Delta t$ (upper bound).
In practice, while the exact value of $\delta$ varies across timesteps and samples, it is inherently bounded by the Lipschitz continuity of the pre-trained network. Furthermore, the upper bound is typically dominated by the inverse step size term, creating a wide margin for feasible $k$.
Consequently, we treat $k$ as a scalar hyperparameter. Our ablation studies (Table 3 in the main paper) empirically verify this theoretical corridor: excessively low $k$ fails to overcome model drift (under-correction), while excessively high $k$ induces numerical chattering (over-correction). A fixed intermediate value provides robust performance across diverse inputs without requiring real-time estimation of $\delta$.

\section{Additional Implementation Details}

\subsection{Datasets and Baselines.}
To comprehensively evaluate the proposed SMC-CFG, we compare it with the standard CFG on the image-generation benchmark T2I-CompBench~\cite{huang2023t2i} using three different flow matching models. T2I-CompBench is a comprehensive benchmark for open-world compositional text-to-image generation, comprising 6,000 compositional text prompts. In our experiments, we focus on four sub-categories that are most relevant to text-aligned image fidelity: color binding, shape binding, texture binding, and spatial relationships. For all flow matching models, we adopt publicly available checkpoints from HuggingFace. Specifically, Stable Diffusion 3.5 is based on the ``stabilityai/stable-diffusion-3.5-large'' public weights. Flux-dev uses ``black-forest-labs/FLUX.1-dev''. Given that Flux-dev is a guidance-distilled model, we set the embedded guidance to 1 in baseline experiments to ensure fairness when no CFG is applied. Qwen-Image uses the ``Qwen/Qwen-Image'' checkpoint. All models generate images at a resolution of 1024 $\times$ 1024 from textual prompts without any additional fine-tuning.

\subsection{Metrics.}
In the main text, we utilize a series of evaluation metrics. FID (Fr\'{e}chet Inception Distance)~\cite{heusel2017FID} computes the Fr\'{e}chet Distance between the multivariate Gaussian distribution estimated from the feature vectors of generated images and of real images, assessing the image quality and diversity of the generated results. CLIP Score~\cite{hessel2021clipscore}, utilizing a pre-trained CLIP~\cite{radford2021clip} model, quantifies the semantic alignment between the generated image and the text prompt by computing the cosine similarity between their respective L2-normalized feature vectors. Aesthetic Score~\cite{schuhmann2022laionAesthetics} serves as an aesthetic regression model, evaluating the image's general aesthetic appeal, such as excellent composition and harmonious coloring. ImageReward~\cite{xu2023imagereward} is a general-purpose reward model trained on a large dataset of expert human preference feedback, which quantifies the generated image's perceived quality and attractiveness to predict the probability of being preferred by humans. PickScore~\cite{kirstain2023pickscore} is a CLIP-based scoring function trained on real users' preference data, specifically designed to predict the probability of a generated image being selected by humans in a competitive setting. HPSv2 and HPSv2.1 (Human Preference Score)~\cite{wu2023HPSv2} are multi-dimensional perceptual metrics that simultaneously assess the image adherence to the prompt, aesthetic quality, and visual fidelity. Finally, MPS (Multi-dimensional Preference Score)~\cite{zhang2024MPS} is a unified model that utilizes a condition mask on top of the CLIP model to predict the quality of a text-to-image output across four distinct human preference dimensions: Overall, Aesthetics, Semantic Alignment, and Detail Quality.

\subsection{Hyperparameters.}

We determine the hyperparameters of SMC-CFG through grid search over the two parameters $\lambda$ and $k$. Specifically, $\lambda$ is searched within $\{2,3,4,5,6,7,8\}$, while $k$ is explored over $\{0.01, 0.05, 0.1, 0.15, ..., 0.75, 0.8\}$. To avoid test-set leakage, the grid search is conducted on an auxiliary set of 200 cases sampled from the MS-COCO~\cite{lin2014mscoco} dataset, which is entirely disjoint from the evaluation set used in the experiment. The optimal configurations selected for the three text-to-image models used in our experiments are as follows: for Stable Diffusion 3.5, $\lambda$ = 6 and $k = 0.1$; for Flux, $\lambda = 6$ and $k = 0.7$; and for Qwen-Image, $\lambda = 6$ and $k = 0.1$. The main experiments adopt these hyperparameter settings without further modification.

\section{More Experiments}

\subsection{Text-to-Image Benchmark Evaluation}

\begin{figure*}[t]
    \centering
    \includegraphics[width=0.85\linewidth]{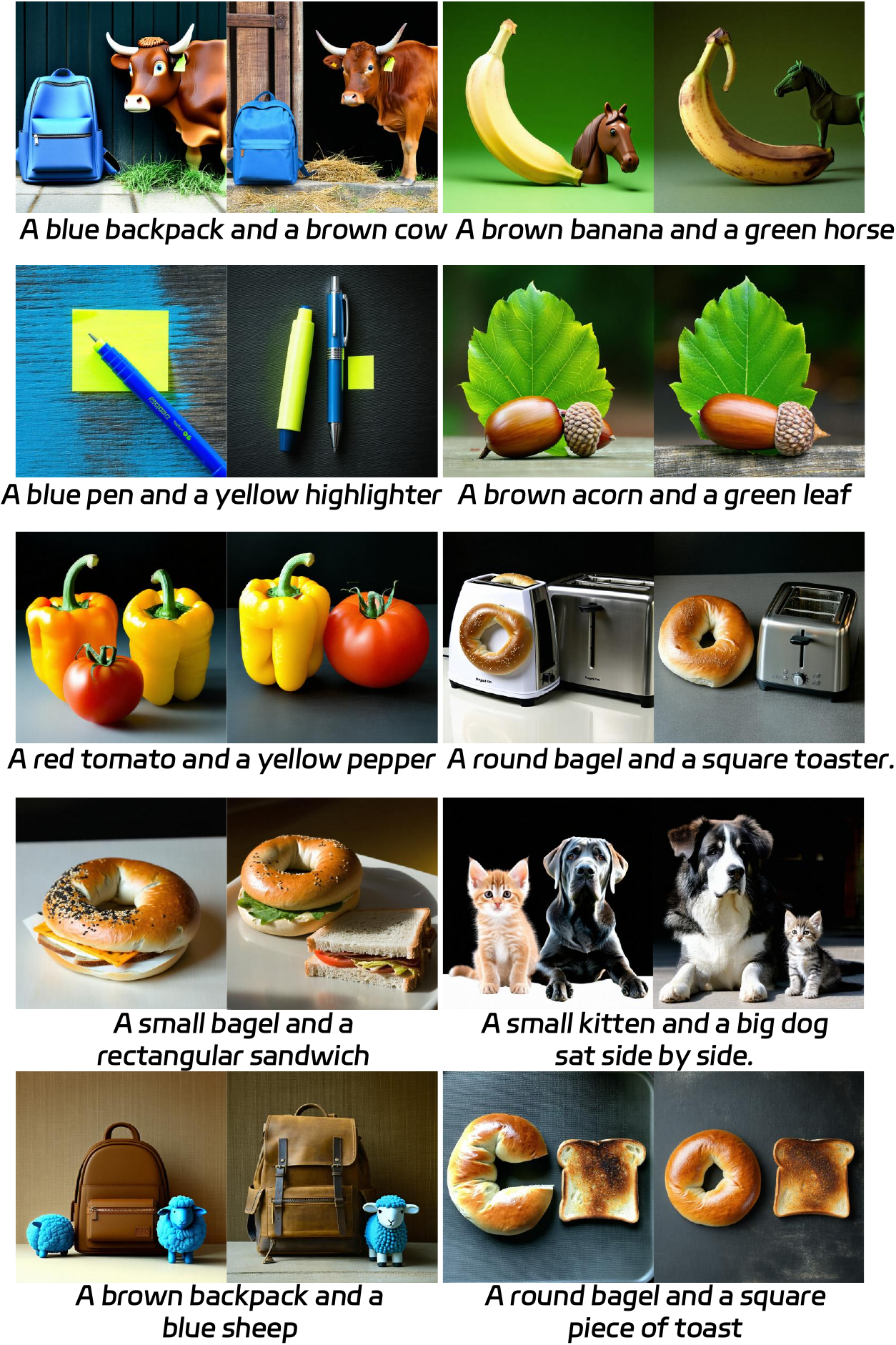}
    \caption{Addtional visual comparison between CFG (left) and SMC-CFG (right) in SD3.5.}
    \label{fig:T2I_compare_sd}
\end{figure*}

\begin{figure*}[t]
    \centering
    \includegraphics[width=0.8\linewidth]{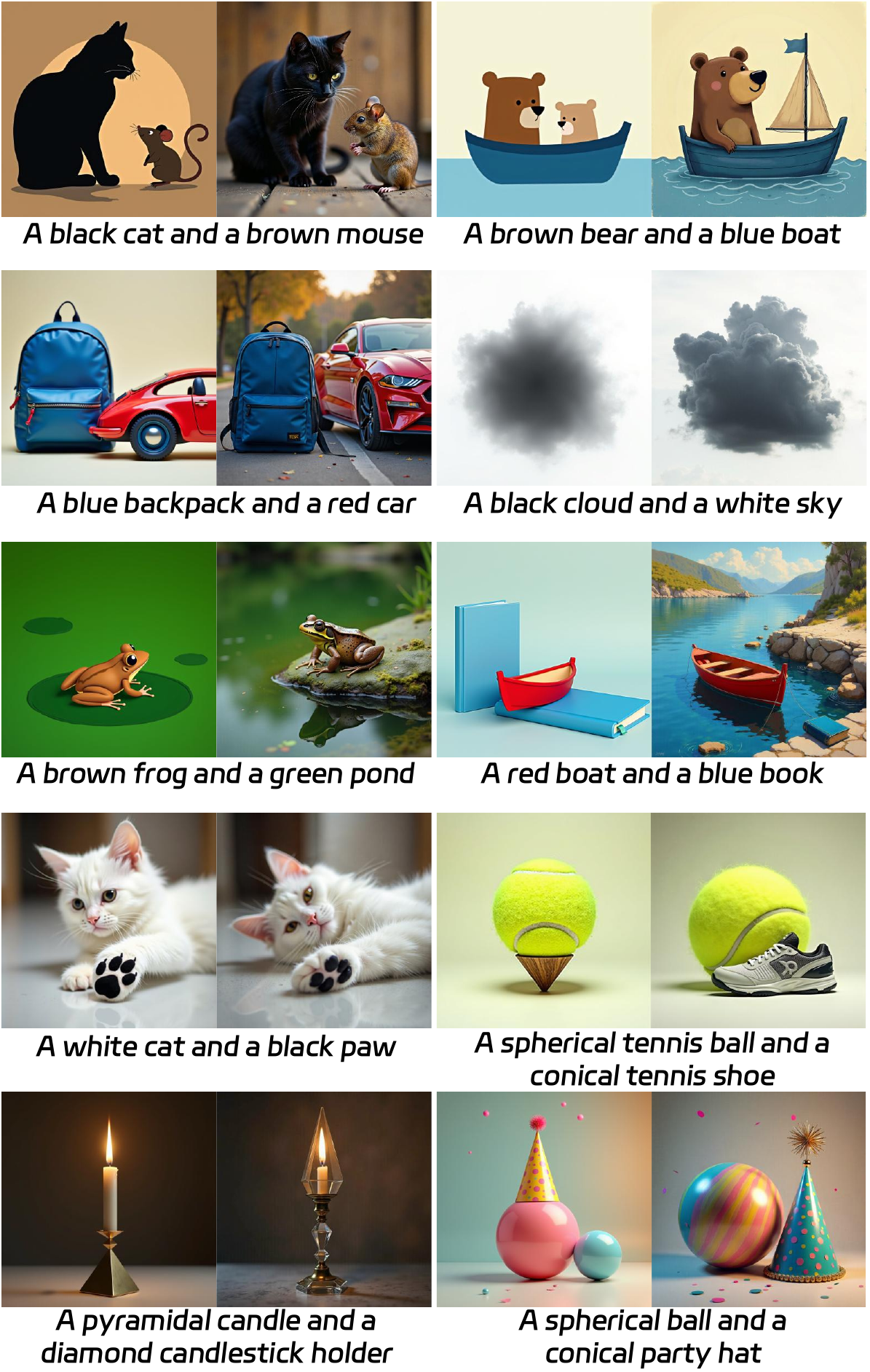}
    \caption{Addtional visual comparison between CFG (left) and SMC-CFG (right) in Flux-dev.}
    \label{fig:T2I_compare_flux}
\end{figure*}

\begin{figure*}[t]
    \centering
    \includegraphics[width=0.8\linewidth]{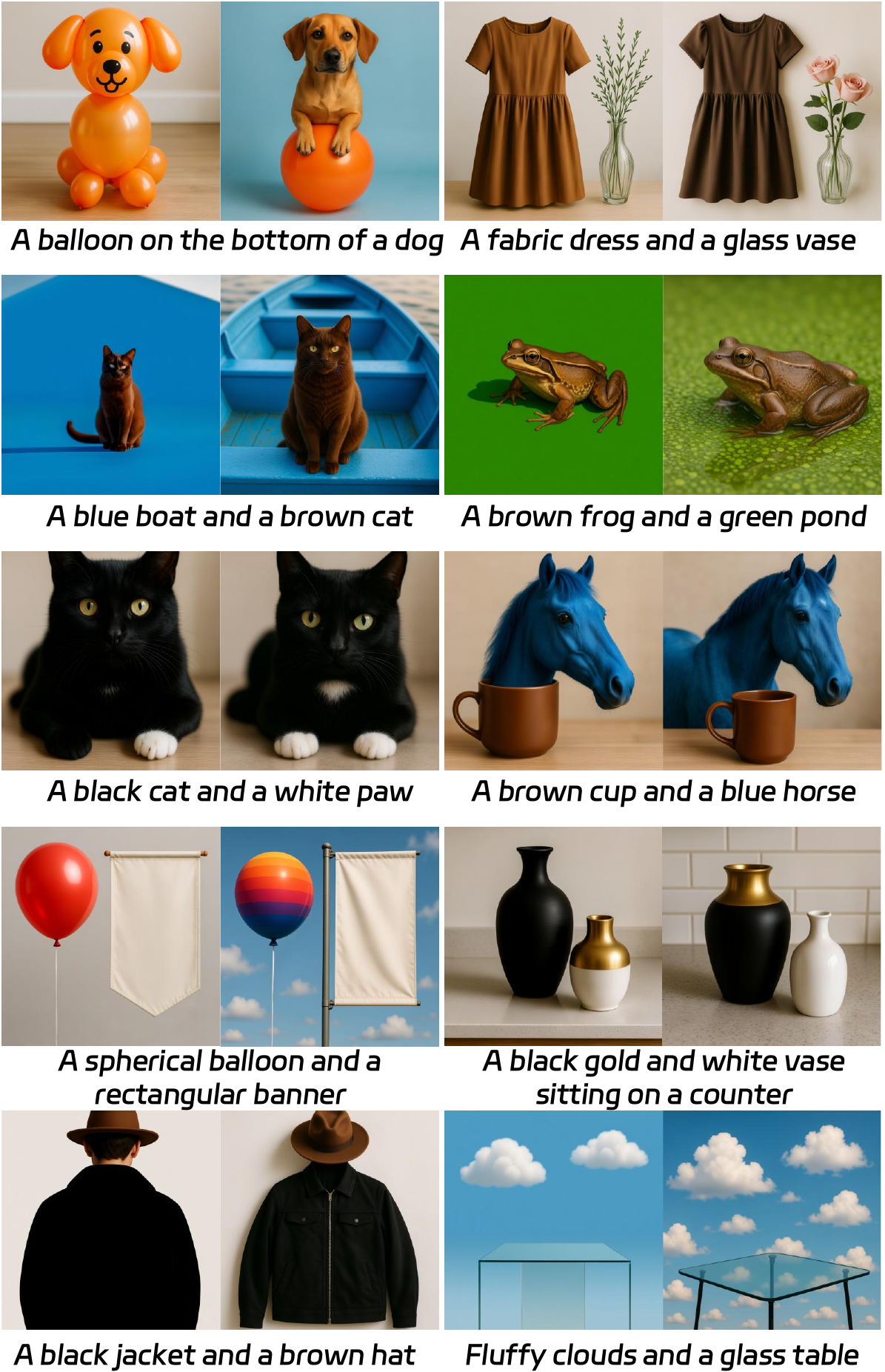}
    \caption{Addtional visual comparison between CFG (left) and SMC-CFG (right) in Qwen-Image.}
    \label{fig:T2I_compare_qwen}
\end{figure*}

We evaluate SMC-CFG on three flow matching text-to-image models using T2I-CompBench~\cite{huang2023t2i}, and compare it with representative CFG-based baselines on VQAScore (GenAI-Bench). Table~\ref{tab:t2i-compbench} shows that SMC-CFG improves the compositional generation performance of SD3.5, Flux-dev, and Qwen-Image on Color, Shape, Texture, and Spatial. The gains are generally larger on spatial and attribute-related dimensions. Table~\ref{tab:VQAScore} reports the VQAScore results on SD3.5. SMC-CFG achieves the best Base, Advance, and Overall scores among the compared methods, outperforming standard CFG as well as recent variants such as CFG-Zero and Rect-CFG++. Visual comparisons on the three T2I models are shown in Figure~\ref{fig:T2I_compare_sd},~\ref{fig:T2I_compare_flux}, and~\ref{fig:T2I_compare_qwen}.

\begin{table}[t]
\centering
\small
\caption{\textbf{Quantitative evaluation on T2I-CompBench.}}
\label{tab:t2i-compbench}
\resizebox{\linewidth}{!}{
    \begin{tabular}{l|cccc}
      \toprule
      \textbf{Model} & \textbf{Color} $\uparrow$ & \textbf{Shape} $\uparrow$ & \textbf{Texture} $\uparrow$ & \textbf{Spatial} $\uparrow$ \\
      \midrule
      SD3.5~\cite{esser2024sd3} & 0.6790 & 0.5915 & 0.7243 & 0.1625 \\
      \textbf{w/ SMC-CFG} & \textbf{0.7461} & \textbf{0.6009} & \textbf{0.7406} & \textbf{0.2563} \\
      \midrule
      Flux-dev~\cite{flux2024} & 0.8172 & 0.5751 & 0.7432 & 0.2708 \\
      \textbf{w/ SMC-CFG} & \textbf{0.8216} & \textbf{0.6199} & \textbf{0.7901} & \textbf{0.2939} \\
      \midrule
      Qwen-Image~\cite{wu2025qwen-image} & 0.7747 & 0.5621 & 0.6747 & 0.2968 \\
      \textbf{w/ SMC-CFG} & \textbf{0.8191} & \textbf{0.5934} & \textbf{0.7421} & \textbf{0.4085} \\
      \bottomrule
    \end{tabular}
}
\end{table}

\begin{table}[ht]
\centering
\small
\caption{\textbf{Compositional alignment evaluation on SD3.5.}}
\label{tab:VQAScore}
\begin{tabular}{l|ccc}
  \toprule
  \multirow{2}{*}{\textbf{Method}} & \multicolumn{3}{c}{\textbf{VQAScore (GenAI-Bench)}} \\
  \cmidrule(lr){2-4}
  & \textbf{Base} $\uparrow$ & \textbf{Advance} $\uparrow$ & \textbf{Overall} $\uparrow$ \\
  \midrule
  Base (w/o CFG) & 0.79 & 0.64 & 0.70 \\
  \midrule
  w/ CFG & 0.83 & 0.64 & 0.72 \\
  w/ CFG-Zero$^\star$ & 0.88 & 0.66 & 0.75 \\
  w/ Rect-CFG++ & 0.87 & 0.64 & 0.73 \\
  \textbf{w/ SMC-CFG} & \textbf{0.89} & \textbf{0.68} & \textbf{0.77} \\
  \bottomrule
\end{tabular}
\end{table}

\subsection{Text-to-Video Generation}
\begin{figure*}[t]
    \centering
    \includegraphics[width=0.8\linewidth]{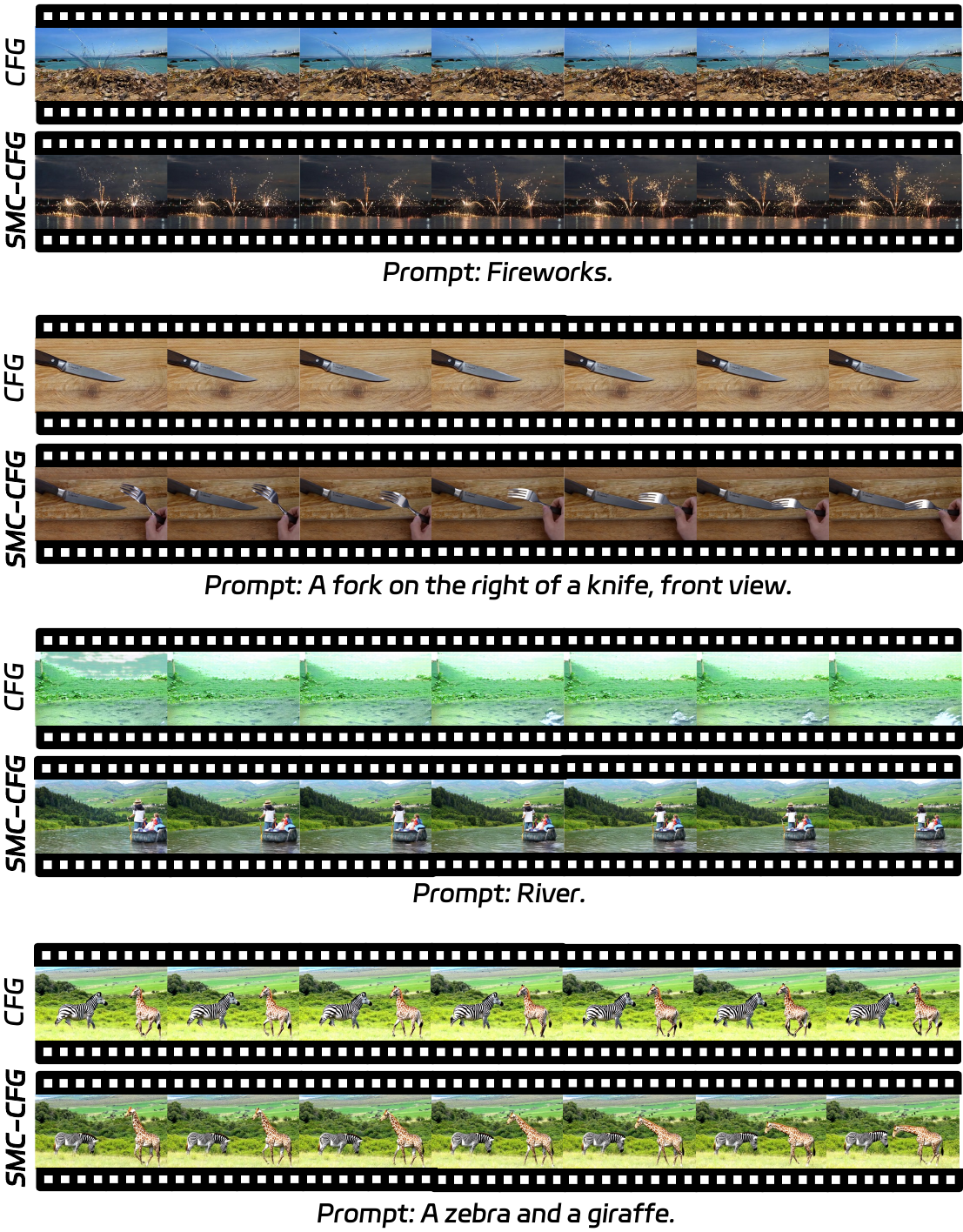}
    \caption{Additional video comparisons between CFG (above) and SMC-CFG (below) in Wan2.2-TI2V-5B.}
    \label{fig:wan_cases}
\end{figure*}

We further extend our evaluation to the text-to-video generation task to assess the generalization capability of SMC-CFG. Using the Wan2.2-TI2V-5B~\cite{wan2025wan} model, we conduct a qualitative comparison against the standard CFG baseline. As visualized in Figure~\ref{fig:wan_cases}, our method demonstrates superior stability in the spatiotemporal domain. Specifically, SMC-CFG enhances temporal consistency, producing smoother motion trajectories with fewer visual artifacts or flickering compared to the baseline. Furthermore, it exhibits robust semantic adherence in complex compositional scenarios, effectively maintaining the spatial structure and identity of generated objects throughout the video sequence. We also show quantitative evaluation in Table~\ref{tab:vbench_comparison}. SMC-CFG improves the total VBench score and gives higher Quality and Semantic scores than CFG. It also performs better on Color, Human Action, and Subject Consistency. These results suggest that the behavior of SMC-CFG is not limited to text-to-image generation and can transfer to text-to-video generation as well.

\begin{table}[t]
\centering
\caption{\textbf{Video comparison on Wan2.2-TI2V-5B.}}
\label{tab:vbench_comparison}
\resizebox{\linewidth}{!}{
    \begin{tabular}{l|cccccc}
      \toprule
      \textbf{Method} & \textbf{Total Score} & \textbf{Quality Score} & \textbf{Semantic Score} & \textbf{Color} & \textbf{Human Action} & \textbf{Subject Consistency} \\
      \midrule
      CFG & 0.5594 & 0.6581 & 0.4607 & 0.9087 & 0.5313 & 0.9450 \\
      \textbf{SMC-CFG} & \textbf{0.5839} & \textbf{0.6747} & \textbf{0.4931} & \textbf{0.9818} & \textbf{0.6000} & \textbf{0.9609} \\
      \bottomrule
    \end{tabular}
}
\end{table}

\subsection{Computational Efficiency}
We further assess the computational overhead and inference latency of our method at different output resolutions to demonstrate its practicality in real-world deployment scenarios. As presented in Table~\ref{tab:comp_cost}, SMC-CFG exhibits memory consumption and FLOPs that are comparable to those of standard CFG in a single inference pass, and the average inference time remains nearly identical. These results indicate that SMC-CFG preserves the computational efficiency of standard CFG and does not introduce additional computational cost or latency during inference.

\begin{table}[ht]
\centering
\captionof{table}{{Computational cost and inference time comparison of standard CFG and SMC-CFG}.}
\label{tab:comp_cost}
\resizebox{\linewidth}{!}{
    \begin{tabular}{l|c|c|c|c}
      \toprule
      \textbf{Resolution} & \textbf{Guidance} & \textbf{Memory (GB)} & \textbf{FLOPs (G)} & \textbf{Runtime (s)} \\
      \midrule
      \multirow{2}{*}{512×512} 
        & CFG & 31.99 & 1203370.06 & 23.84 \\
        & \textbf{SMC-CFG} & 31.99 & 1203370.07 & 23.97 \\
      \midrule
      \multirow{2}{*}{1024×1024}
        & CFG & 33.59 & 3590870.89   & 44.78 \\
        & \textbf{SMC-CFG} & 33.59 & 3590870.93 & 45.09 \\
      \bottomrule
    \end{tabular}
}
\end{table}

\subsection{Ablation Study on Hyperparameter Effects}

\begin{figure}[h]
    \centering
    \includegraphics[width=0.95\linewidth]{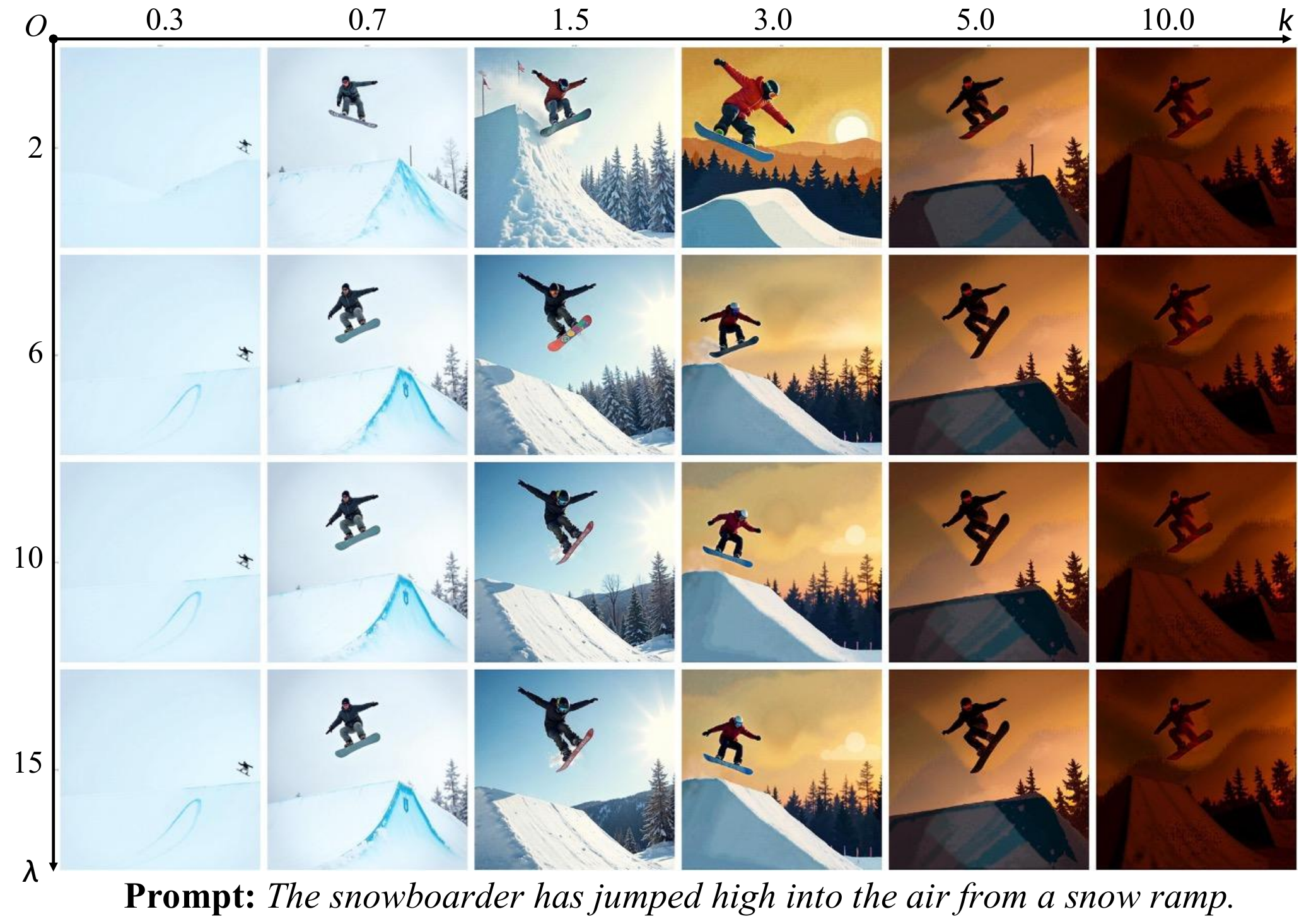}
    \caption{Qualitative results under various hyperparameters.}
    \label{fig:ablation}
\end{figure}

We conduct visual comparison with fixed initial noise to show impact of hyperparameters. As shown in Figure~\ref{fig:ablation}, $\lambda$ governs the stability of structural details by shaping the sliding mode manifold, while $k$ regulates the overall semantic alignment and its trade-off with aesthetic realism.

\section{More Discussion}
\subsection{CFG Scale}
\begin{figure}[ht]
    \centering
    \includegraphics[width=\linewidth]{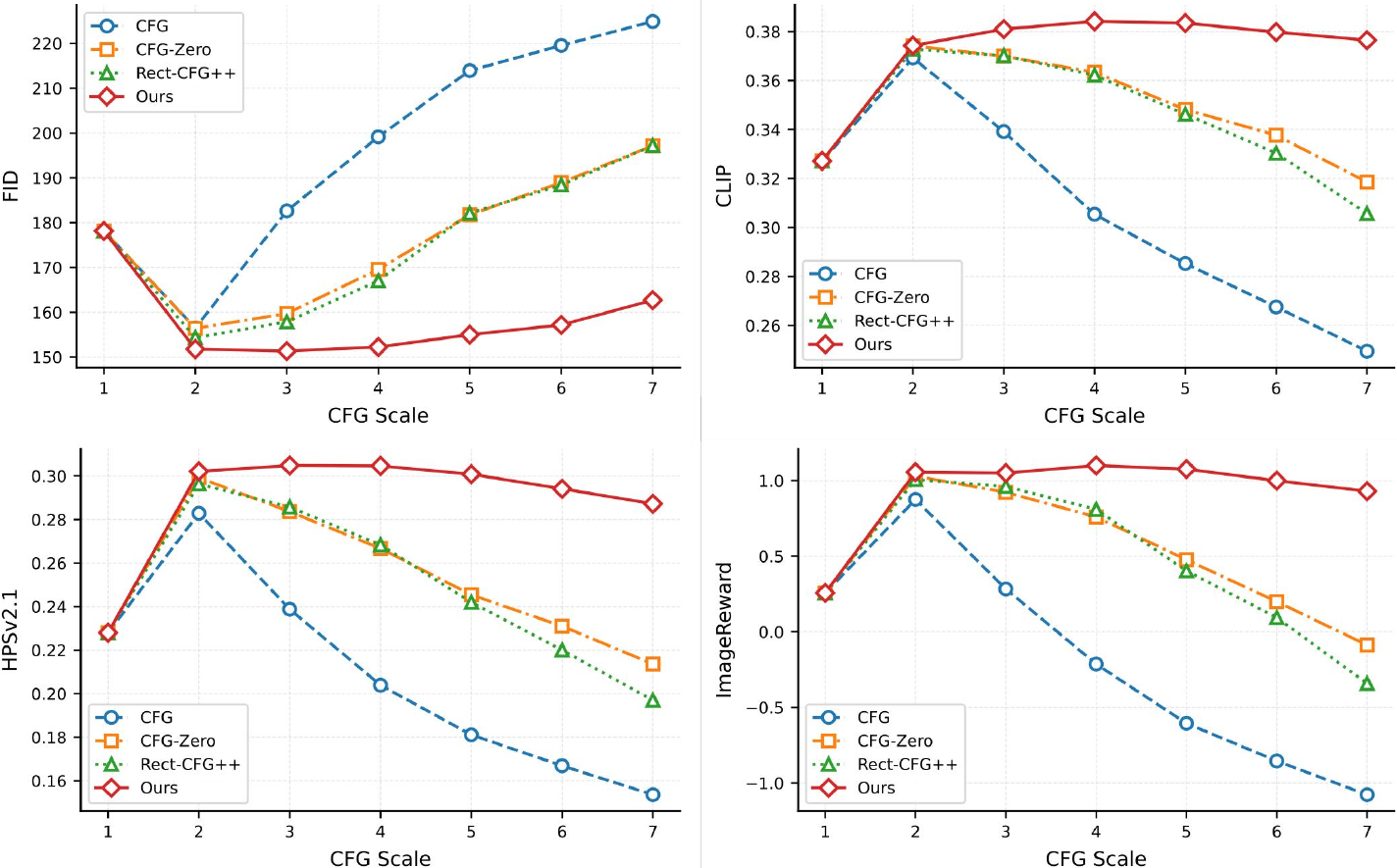}
    \caption{Performance curves of different methods under varying CFG scales.}
    \label{fig:cfg_scale_curve}
\end{figure}

We analyze the effect of the CFG scale by visualizing the performance curves of the main evaluation metrics under varying guidance strengths on the Flux-dev model, as shown in Figure~\ref{fig:cfg_scale_curve}. When the CFG scale reaches the model’s default optimal value of 2, both standard CFG and other baseline methods achieve their best performance. However, their performance rapidly degrades as the scale increases further, revealing the strong nonlinear distortions introduced by high guidance. In contrast, SMC-CFG continues to improve as the CFG scale increases, demonstrating that it can better exploit the potential of guidance without suffering from the instability observed in conventional methods. Even under extremely large scales, SMC-CFG shows only a slight performance drop, indicating strong robustness against over-guidance effects.

\subsection{Limitations and Future Work}
Despite its ability to alleviate the nonlinear effects associated with high CFG scales and to substantially improve compositional image generation, SMC-CFG introduces two additional hyperparameters, which increase the complexity of deployment and may require manual tuning for different models. In the future, we plan to explore adaptive guidance control mechanisms capable of dynamically adjusting control parameters according to the evolving state of the generative process. In particular, one promising way is to incorporate error-differential feedback, where changes in text–image alignment across successive steps are used to automatically increase or decrease the effective guidance strength. The adaptive strategy offers the potential to eliminate manual tuning while improving stability and performance under varying guidance scales.

\end{document}